%% file: main.tex
\documentclass[lettersize,journal]{IEEEtran}
\usepackage[dvipsnames,table,xcdraw]{xcolor}
\usepackage[pagebackref=true,breaklinks=true,colorlinks,citecolor=green]{hyperref}         
\usepackage{amsmath,amsfonts}
\usepackage{algorithmic}
\usepackage{algorithm}
\usepackage{array}
\usepackage[caption=false,font=normalsize,labelfont=sf,textfont=sf]{subfig}
\usepackage{textcomp}
\usepackage{stfloats}
\usepackage{url}
\usepackage{verbatim}
\usepackage{cite}
\usepackage{times}
\usepackage{epsfig}
\usepackage{graphicx}
\usepackage{amsmath}
\usepackage{amssymb}
\usepackage{booktabs}
\usepackage{caption}
\usepackage{multirow}
\usepackage[accsupp]{axessibility} 
\usepackage{xr}
\usepackage{indentfirst}
\usepackage{everyshi}
\usepackage{cleveref}
\usepackage[misc]{ifsym}
\hypersetup{
  pdftitle={IntentQA: Intent Question Answering in Videos by Cognitive Context Reasoning},
  pdfauthor={Jiapeng Li, Ping Wei, Wenjuan Han, Song-Chun Zhu, and Lifeng Fan},
  pdfsubject={Accepted manuscript of an article published in IEEE Transactions on Pattern Analysis and Machine Intelligence},
  pdfkeywords={intent understanding, social intelligence, video question answering, visual reasoning, context}
}

\begin{document}
     
\title{IntentQA: Intent Question Answering in Videos \\ by Cognitive Context Reasoning}

\author{Jiapeng Li$^{1}$,
        Ping Wei$^{\textrm{\Letter},1}$,~\IEEEmembership{Senior Member,~IEEE} \\
        Wenjuan Han$^{3}$,
        Song-Chun Zhu$^{2}$,~\IEEEmembership{Fellow,~IEEE,}
        Lifeng Fan$^{\textrm{\Letter},2}$
\thanks{\footnotesize\noindent

$^1$ National Key Laboratory of Human-Machine Hybrid Augmented Intelligence, Xi'an Jiaotong University, Xi'an, China

$^2$ National Key Laboratory of General Artificial Intelligence, Beijing Institute for General Artificial Intelligence (BIGAI), Beijing, China 

$^3$ School of Computer and Information Technology, Beijing Jiaotong University, Beijing, China

$\textrm{\Letter}$ Corresponding authors: Ping Wei (pingwei@xjtu.edu.cn) and Lifeng Fan (lifengfan@bigai.ai).
}
\thanks{Manuscript received 13 July 2024; revised 21 November 2025; accepted 26 April 2026.}
\thanks{\copyright~2026 IEEE. Personal use of this material is permitted. Permission from IEEE must be obtained for all other uses, in any current or future media, including reprinting/republishing this material for advertising or promotional purposes, creating new collective works, for resale or redistribution to servers or lists, or reuse of any copyrighted component of this work in other works.}
\thanks{This is the accepted manuscript. The Version of Record is: J. Li, P. Wei, W. Han, S.-C. Zhu, and L. Fan, ``IntentQA: Intent Question Answering in Videos by Cognitive Context Reasoning,'' \textit{IEEE Transactions on Pattern Analysis and Machine Intelligence}, vol. 48, no. 9, pp. 11044--11061, September 2026, doi: \href{https://doi.org/10.1109/TPAMI.2026.3690561}{10.1109/TPAMI.2026.3690561}.}}

\markboth{IEEE Transactions on Pattern Analysis and Machine Intelligence,~Vol.~48, No.~9, September~2026}%
{Li \MakeLowercase{\textit{et al.}}: IntentQA: Intent Question Answering in Videos by Cognitive Context Reasoning}

\Crefname{figure}{Fig.}{Figs.}%
\Crefname{equation}{Eq.}{Eqs.}%

\maketitle

\begin{abstract}
\textcolor{black}{Video understanding requires intelligent agents to transcend mere recognition of visual facts and comprehend the underlying intents behind human actions—often termed the ``dark matter'' of social intelligence. To bridge the gap between visual observation and intent reasoning,} we introduce a novel task, \textbf{IntentQA}, \textcolor{black}{and contribute a large-scale VideoQA dataset specifically tailored for this purpose.}
\textcolor{black}{However, recognizing that standard metrics may overestimate capabilities due to dataset biases, we go beyond simple accuracy to rigorously evaluate model robustness. We augment the benchmark by generating five distinct contrast sets via Large Language Models (LLMs) and introducing a ``Contrast Performance Decline'' metric.}
We propose the \textcolor{black}{\textbf{X-CaVIR} (eXplainable Context-aware Video Intent Reasoning)} framework, which leverages \textcolor{black}{three types of} ``Cognitive Context'' to enhance video analysis: i) \textcolor{black}{Situational Context via a cross-modal} Video Query Language (VQL) \textcolor{black}{module}, ii) \textcolor{black}{Contrastive Context via a} Contrastive Learning module, and iii) \textcolor{black}{Commonsense Context via a} Commonsense Reasoning module.
\textcolor{black}{Crucially, to overcome the opacity of traditional black-box models, we refine the integration of LLMs within X-CaVIR by employing a transparent pipeline that synergizes video captions with VQA model outputs. This approach not only improves performance by effectively utilizing rich commonsense knowledge but also renders the reasoning process explicitly interpretable.}
Extensive experiments demonstrate the effectiveness of \textcolor{black}{our components}, the superiority of \textcolor{black}{X-CaVIR} over \textcolor{black}{state-of-the-art} baselines, \textcolor{black}{and its stability against perturbations on the contrast sets.} The dataset and codes are open-sourced at: \url{https://github.com/JoseponLee/IntentQA}.
\end{abstract}


\section{Introduction}
\label{sec:intro}

\begin{figure*}[!t]
\begin{center}
   \includegraphics[width=1\linewidth]{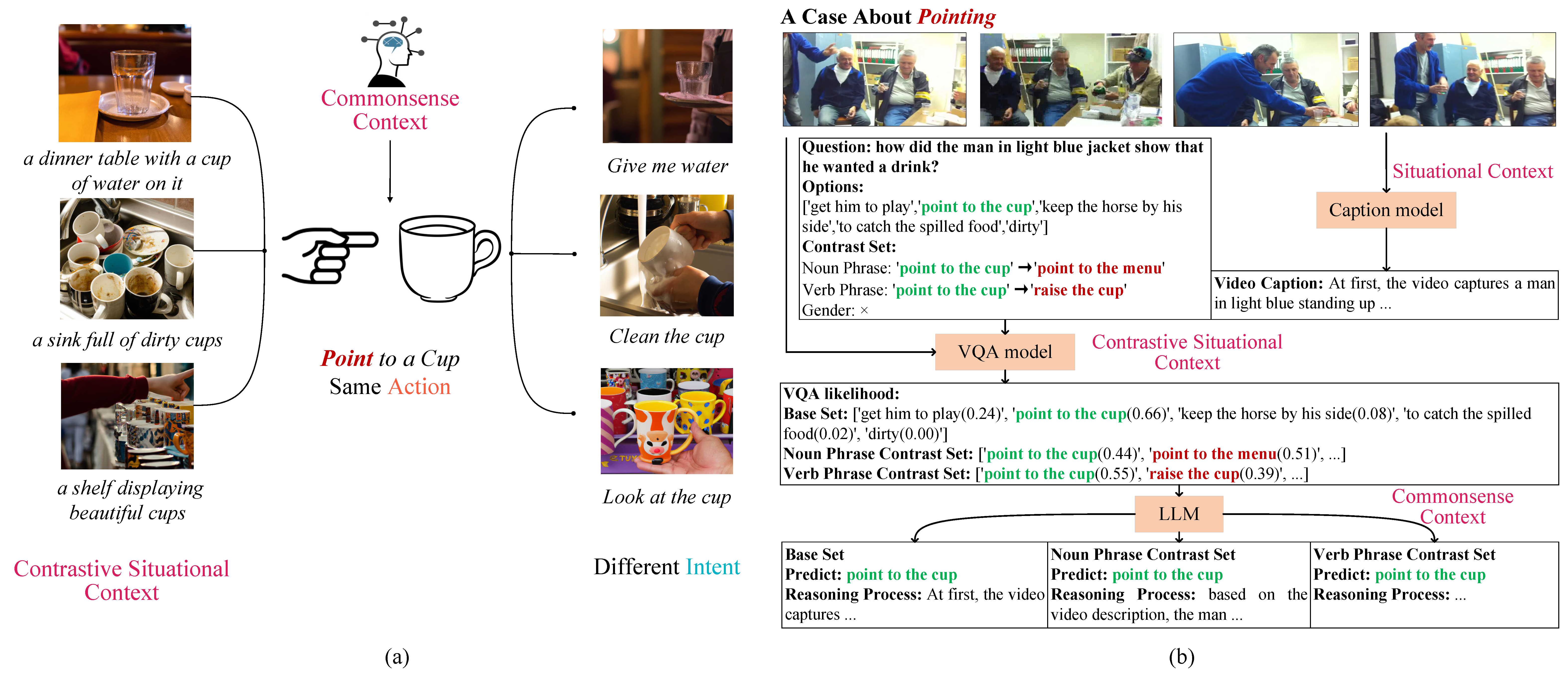}
\end{center}
\vspace{-0.4cm}
   \caption{(a) Illustration of challenges brought by varied contexts in video intention reasoning. The same action under different contexts could mean different underlying intents. (b) Illustration of the overview of our model's inference and an example of its contrast set.}
\label{fig:intro}
\vspace{-0.4cm}
\end{figure*}

\textcolor{black}{Video question answering (VideoQA) has emerged as a prominent field for interactive AI, yet it faces significant challenges in moving beyond factoid-based tasks to inferential reasoning about latent variables \cite{zhong2022video}.} 
\textcolor{black}{While current models excel at recognizing observable facts, they struggle with the ``dark matter'' of social intelligence—the unobservable mental states that drive human behavior \cite{zhu2020dark, fan2022artificial}.} 
Psychological research demonstrates that humans are adept at inferring this mental world from simple visual cues \cite{heider1944experimental}. 
Even six-month-old infants perceive human actions as goal-directed behavior \cite{woodward1998infants}, and adults naturally interpret events as sequences of goals driven by intentions \cite{csibra2007obsessed, WeiCVPR18HAO, WeiIJCAI17LInt}. 
In fact, rather than encoding every action detail, humans store interpretations of intent for later retrieval \cite{baldwin2001discerning}. 
\textcolor{black}{However, bridging the gap between visual observation and intent reasoning remains essential for building truly intelligent systems.}

To address this, we define the task of \textbf{IntentQA}, which requires an AI agent to answer questions regarding the psychological goals and distinct intentions of characters in a video. We argue that successful intent reasoning relies heavily on three distinct types of cognitive contexts:

(I) \textbf{Situational Context}: 
As illustrated in \Cref{fig:intro} (a), this pertains to the immediate environment in which the action occurs, emphasizing how contextual elements directly influence the interpretation. For example, the action of ``pointing to a cup'' is interpreted as ``give me water'' in a dining context, but implies ``clean the cup'' when the context is ``a sink full of dirty cups.'' The immediate surroundings play a crucial role in shaping the understanding of non-verbal cues.

(II) \textbf{Contrastive Context}: 
This involves comparing the observed action with similar actions in different scenarios to highlight distinct intents. For instance, by contrasting the scenario of ``a sink full of dirty cups'' against ``a shelf displaying beautiful cups,'' the divergent meanings attached to the same pointing action (i.e., cleaning vs. admiring) become distinguishable. This comparison facilitates a deeper, context-sensitive understanding that enhances adaptability.

(III) \textbf{Commonsense Context}: 
This context relies on intrinsic or culturally ingrained understandings that link specific scenes with typical actions. For example, placing a cup on a table in a library is commonly understood as reserving a seat. This shared cultural knowledge bridges the gap between an observed action and its underlying purpose, enabling the inference of intent from seemingly simple actions that would otherwise appear to have no deeper meaning.

We contribute a new dataset for IntentQA, organized into four categories: \textit{Causal Why}, \textit{Causal How}, \textit{Temporal Previous}, and \textit{Temporal Next}. 
\textcolor{black}{However, recent critiques suggest that standard evaluation metrics may overestimate model capabilities due to dataset biases \cite{gardner2020evaluating, park2022exposing}. Specifically, Gardner \textit{et al.} \cite{gardner2020evaluating} demonstrated that popular models often suffer significant performance drops when tested on ``contrast sets'' designed with subtle perturbations.}
\textcolor{black}{To address this, we propose \textbf{X-CaVIR} (eXplainable Context-aware Video Intent Reasoning), a framework that utilizes Large Language Models (LLMs) as explicit reasoners. By incorporating a novel pipeline that synergizes video captions with VQA confidence scores, X-CaVIR achieves superior performance while providing an observable window into the reasoning process.}

\textbf{Our contributions} are summarized as follows:
1) We introduce IntentQA, a novel VideoQA task focusing on intent reasoning, where the objective is to accurately select answers based on understanding of intents.
2) We collect and annotate an original large-scale IntentQA dataset. \textcolor{black}{Crucially, we construct a rigorous robustness benchmark by generating five distinct contrast sets and introducing the ``Contrast Performance Decline'' metric.}
3) We build a Context-aware Video Intent Reasoning model (\textcolor{black}{X-CaVIR}) that incorporates Situational, Contrastive, and Commonsense contexts.
4) \textcolor{black}{We refine the integration with LLMs in X-CaVIR, employing a transparent pipeline that yields superior performance while offering interpretable reasoning pathways.}

A preliminary version of this work \cite{Li2023IntentQA} has been accepted for oral presentation at ICCV 2023. \textcolor{black}{The current version presents several significant advancements over the previous one:}
\begin{itemize}
    \item \textcolor{black}{\textbf{From Accuracy to Robustness:} The original task definition relies on standard accuracy, which can be prone to dataset biases. In this extension, we introduce a comprehensive robustness evaluation framework by generating five distinct Contrast Sets (involving verb, noun, and gender perturbations) and a new metric, Contrast Performance Decline. This ensures that the model's performance reflects genuine intent understanding rather than superficial pattern matching.}
    \item \textcolor{black}{\textbf{From Black-box to Interpretable Reasoning:} While the preliminary model (CaVIR) successfully introduced context awareness, its reasoning mechanism remained largely opaque. In this version, we propose X-CaVIR, which features a novel pipeline integrating VideoQA confidence scores with dense captions and LLMs. This upgrade not only significantly improves performance but also provides an interpretable reasoning process, allowing us to explicitly trace why an intent is inferred.}
\end{itemize}

\textcolor{black}{The rest of the paper is organized as follows. Section~\ref{related} reviews related work. Section~\ref{sec:Dataset} presents the proposed IntentQA dataset in detail. Section~\ref{sec:model} describes the proposed X-CaVIR model. Section~\ref{sec:experiment} demonstrates the effectiveness of our method through extensive experiments.}

\section{Related Work}\label{related}

\subsection{Video Question Answering}

As a typical cross-modal task, VideoQA answers the natural language question according to the given video, which is challenging because it requires a deep and comprehensive understanding of the semantic information of the video and question. Notably, recent studies in this domain have shifted away from the traditional reliance on 3D convolutions \cite{carreira2017quo, li2020slow} as the primary video backbone models. Instead, approaches harnessing fine-grained information, such as objects and relations, are increasingly gaining traction \cite{xiao2022video, xiao2023contrastive}. A growing body of work recognizes the paramount importance of `context' in addressing problem. 

On the one hand, VideoQA datasets evolve over time \cite{zhong2022video}. In addition to the early datasets, such as TGIFQA \cite{jang2017tgif} and MSRVTT-QA \cite{xu2017video}, many more challenging datasets have emerged recently, such as NExT-QA \cite{xiao2021next}, CLEVRER \cite{yi2020clevrer}, CLEVR\underline{\hspace{0.5em}}HYP \cite{sampat2021clevr_hyp}, AGQA 2.0 \cite{grunde2022agqa} and Causal-VidQA \cite{li2022representation}, which usually invoke complicated spatial, temporal and causal inference among multiple entities and relations \cite{zhong2022video}. On the other hand, various techniques have been developed for VideoQA \cite{sun2021video, zhong2022video}, such as Memory \cite{fan2019heterogeneous, tapaswi2016movieqa}, Attention \cite{zeng2017leveraging, zhao2017video}, Transformer \cite{xiao2022video, yang2021just}, Neural Modular Networks \cite{le2020hierarchical, qian2022dynamic}, Neural-Symbolic methods \cite{yi2020clevrer, chen2021grounding, ding2021dynamic}, and Graph-structured methods \cite{xiao2022videoas, xiao2022video}.

The inspiring and promising trend from recognition to reasoning in the field of VideoQA is great progress. Answering questions like ``what'' is no longer the core of VideoQA, we further want to answer questions like ``why'' and ``how''. However, although there are studies aiming to reason about various relationships between the visual facts (e.g., \cite{liang2022visual, park2022exposing}), few VideoQA work studies the unobserved human mental state underlying the apparent entities. To our best knowledge, our study is the first VideoQA work focusing on ``intent''. We believe intent-related VideoQA features human-level in-depth understanding of videos, demands higher-level reasoning abilities, and would promote VideoQA toward the core of human intelligence.

\subsection{Intent Understanding}
    
Upon seeing human actions, humans have an inherent tendency to infer other people's intentions from their actions \cite{blakemore2001perception}. Intent understanding plays a key role in human social intelligence \cite{zhu2020dark, fan2022artificial, peng2023tong, QiPAMI21GEP}. There have been some studies exploring intent inference in computer vision, robotics, etc. Jia \textit{et al.} \cite{jia2021intentonomy} collected an social media image dataset \textit{Intentonomy} with an aim to analyze how visual information can facilitate recognition of human intent. Pei \textit{et al.} \cite{pei2011parsing} inferred the goals and intents of agents through an event parsing algorithm. Some studies \cite{social_force2009, neverwalkalone, tang2022evostgat, shu2018perception} manifest human intentions by predicting their trajectories. Holtzen \textit{et al.} \cite{holtzen2016inferring} proposed a method for robots to infer a human’s hierarchical intent from partially observed RGBD videos. Yu-Ching \textit{et al.} \cite{chiu4013382enhancing} used a QA approach in robotic systems to construct interactive dialogue systems, assisting robots in understanding user intentions. Sap \textit{et al.} \cite{sap2022neural} measured the large language model's ability to understand intents and reactions of participants in social interactions. However, there has not yet been a cross-modal intent reasoning video dataset nor a benchmark model in VideoQA. Chen \textit{et al.}~\cite{chen2025mllms} proposed an efficient multimodal interpretable method for explaining MLLM VQA task.

\subsection{Context-aware Reasoning}

Context, including not only the immediate context in videos and languages but also the commonsense knowledge, is very important for answering inference questions because knowledge underpins reasoning \cite{liu2022rainier, zheng2021grice, yuan2020joint}. Research has demonstrated that when relevant knowledge is provided as additional context to commonsense question answering, it can substantially enhance performance \cite{liu2022rainier,chen2025generalized}. Many methods that utilize objects in images to integrate context have been proposed \cite{wan2021unsupervised,lou2022unsupervised, FanCVPR18SAtt, fan2019understanding, fan2021learning, chen2025interpreting}. Zheng \textit{et al.} \cite{zheng2019intention} proposed a novel approach for generating image captions with guiding objects. 
AI continues to be narrow and brittle due to its lack of reasoning ability of context, such as commonsense \cite{choi2022curious}. 
Recent years have brought about a renewed interest in commonsense representation and reasoning \cite{hwang2021comet, huang2019cosmos, sap2019socialiqa, fang2020video2commonsense, zellers2019recognition, chen2024less, chen2025less}. Current systems either rely on external knowledge bases (KBs) to incorporate additional relevant knowledge, or resort to pre-trained language models as the sole implicit source of world knowledge \cite{shwartz2020unsupervised, bosselut2021dynamic}. Hwang \textit{et al.} \cite{hwang2021comet} built a new commonsense knowledge graph, ATOMIC2020. Lourie \textit{et al.} \cite{lourie2021unicorn} argued that QA-based commonsense datasets transfer well with each other, while commonsense knowledge graphs do not. Rainier \cite{liu2022rainier} learns to generate contextually relevant knowledge in response to given commonsense questions. Arabshahi \textit{et al.} \cite{arabshahi2021conversational} used a transformer-based generative commonsense knowledge base as its source of background knowledge for reasoning. In contrast to crowdsourcing, a pre-trained language model like GPT \cite{ouyang2022training} is a more flexible source of external knowledge and a better way to generate large-scale dialogue datasets with social commonsense knowledge, such as SODA \cite{kim2022soda}. West \textit{et al.} \cite{west2021symbolic} show how to selectively distill high-quality causal commonsense from GPT-3. Liu \textit{et al.} \cite{liu2021generated} used external knowledge generated from a language model to improve model performance on four commonsense reasoning tasks. Furthermore, attribution-based interpretable methods~\cite{chen2025interpreting} can also effectively explain the model's reasoning process.

\subsection{Integrating LLMs in V-L Models}

The burgeoning field of Visual-language (V-L) models has increasingly leveraged Large Language Models (LLMs) to enhance performance. Recent integrated approaches typically involve open-source LLMs from the LLAMA series \cite{touvron2023llama, zheng2023judging}, an open-source visual encoder \cite{li2022blip, li2023blip, radford2021learning}, and a linear layer for visual-language alignment. These methods often introduce new datasets for instruct tuning the alignment layer while keeping the core model weights static. Despite their more unified structure and additional training, these integrated models \cite{liu2023visual, zhu2023minigpt, li2023videochat, maaz2023video} do not necessarily outperform system-based methods. This performance gap is primarily due to the inferiority of the LLAMA series \cite{touvron2023llama, zheng2023judging} compared to proprietary models like GPT-4 or ChatGPT \cite{openai2023gpt4, ouyang2022training}. Additionally, the integrated models suffer from poor interpretability, as the specific information obtained by the visual component remains unclear, making it difficult to pinpoint issues within the model.

Early efforts in the VQA domain, such as PICa \cite{yang2022empirical}, KAT \cite{gui2021kat}, REVIVE \cite{lin2022revive}, and Prophet \cite{shao2023prompting}, explored more straightforward integrations of LLMs with vision models. PICa proposed using GPT-3 as a reasoner, utilizing captions to bridge visual inputs and the linguistic capabilities of LLMs. In contrast, KAT and REVIVE focused on using LLMs to gather evidence from captions and QAs, which were then fed into a VQA model for answer selection. Prophet combined initial VQA model predictions with LLM reasoning for final decision-making. These methods employed LLMs either as reasoners or as assistants providing supplementary information.

More recent advancements have seen the development of system-based methods, which integrate LLMs with multiple vision models for more complex tasks. These methods, exemplified by models such as MM-REACT \cite{yang2023mm} and ViperGPT \cite{suris2023vipergpt}, involve briefing the LLM on available APIs and their functionalities, after which the LLM decides which models to use. MM-REACT employs iterative self-reflection by the LLM to assess task requirements and determine necessary actions or additional models. ViperGPT leverages the Chain of Thought \cite{wei2022chain} process to generate executable code for tackling complex tasks. These methods represent a more sophisticated evolution of earlier approaches, with the LLM serving as a central reasoner coordinating various vision models.

Our work is inspired by these advancements, particularly the system-based methods that position the LLM as a reasoner. We adopt this approach, further requiring the LLM to provide detailed reasoning processes. This capability has significantly aided our qualitative analysis of failure cases, enabling us to identify model weaknesses and improve performance. By leveraging the LLM's reasoning capabilities, we gain deeper insights into the model's decision-making process, which is crucial for refining and enhancing our V-L model.

\section{Dataset}
\label{sec:Dataset}
\begin{figure}[t]
\begin{center}
\includegraphics[width=0.5\textwidth]{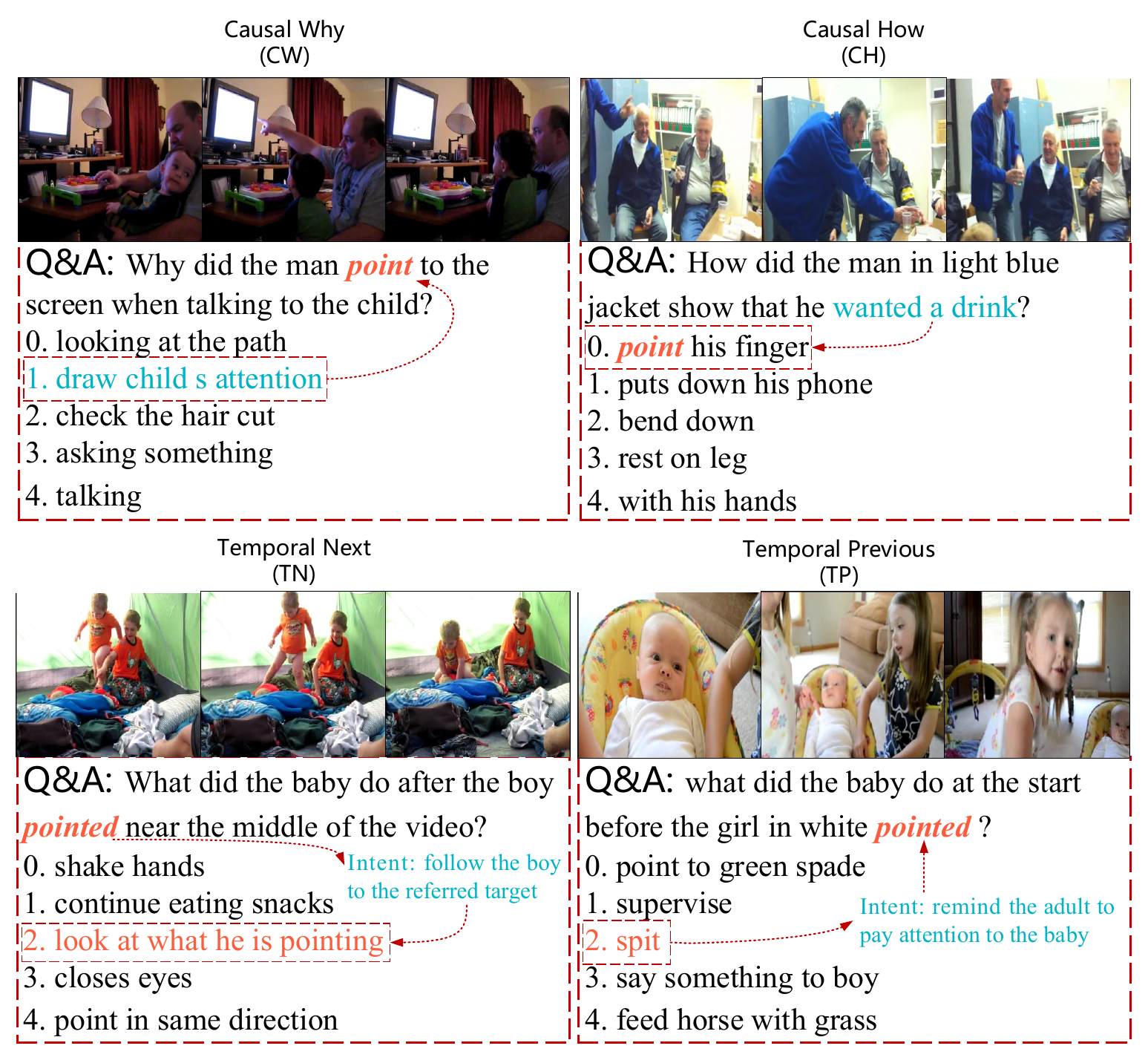}
\end{center}
\vspace{-0.4cm}
\caption{Illustration of four types of QA in our dataset. In the example of \textit{CW}, a man points to the screen to draw the child's attention. In the example of \textit{CH}, the man points his finger to express his intent for a drink. The \textit{TN} example shows that the boy's ``pointing'' leads to the baby's ``looking at what he is pointing''. The \textit{TP} example shows that the girl's ``pointing'' is motivated by the baby's ``spit'' action. The red box frames the correct answer.}
\label{fig:dataset-example}
\end{figure}

We contribute an IntentQA dataset with diverse intents in daily social activities. Examples are shown in \Cref{fig:dataset-example}. We first annotate the main set of the IntentQA dataset, and then extend it with a contrast set that includes several contrast subsets. Next, we introduce the process of dataset construction and annotation for the main set.

\subsection{IntentQA Dataset Part I: the Main Set}
\textbf{Dataset Construction and Annotation.}
In constructing the main set of our dataset, we utilize NExT-QA \cite{xiao2021next} as the foundational dataset. NExT-QA dataset is a comprehensive VideoQA dataset with rich natural daily social activities and detailed QA annotations. Originally, the NExT-QA dataset categorizes itself into three types, i.e., \textit{Causal}, \textit{Temporal}, \textit{Descriptive}. We select the inference QA types, i.e., \textit{Causal} and \textit{Temporal}, rather than the factoid \textit{Descriptive}, to build our IntentQA dataset. Particularly, we select both the \textit{Causal Why} and \textit{Causal How} subtypes under \textit{Causal}, and the \textit{Temporal Previous} and \textit{Temporal Next} subtypes under \textit{Temporal} (see examples shown in \Cref{fig:dataset-example}). The \textit{Causal Why (CW)} QA usually takes the form of ``Why [action]? For [intent]'', with the key action appearing in the question and the intent in the answer. On the contrary, the \textit{Causal How (CH)} QA usually takes the form of ``How [intent]? By [action]'', with the key action appearing in the answer and the intent in the question. The \textit{Temporal Previous (TP)} QA usually takes the form of ``What [action A] before [action B]? '', while the \textit{Temporal Next (TN)} QA takes the form of ``What [action B] after [action A]? '' In the \textit{TP\&TN} QA, the intent is not explicitly expressed in the question nor answer, but is the implicit causal factor linking the two sequential actions. 

We use AllenNLP \cite{Dozat2017DeepBA} for dependency parsing to extract the key action in QA, and obtain the lemmatized verb \footnote{Wordnet:~\url{https://www.nltk.org/_modules/nltk/stem/wordnet.html}} of the action from the dictionary \cite{miller1995wordnet, miller1998wordnet}. We searched for synonyms based on each action's lemmatized verb, and merge the synonyms to assign an action ID for each cluster. After the preliminary filtering and processing, we further annotate the dataset on Amazon Mechanical Turk (AMT). We carefully design four questions to select QAs satisfying the following criteria: 1) The key action is physical, observable in the video, and conducted by a person; 2) The \textit{same actions} refer to semantically the same and physically similar actions in the videos, rather than different actions under the same or similar action words. We construct our dataset in a contrastive manner that the same actions under different contexts lead to different underlying intents, as illustrated in \Cref{fig:intro} (a). To ensure the annotation quality, we apply the cross-validation principle and assign at least three annotators for each data sample; only when all three annotators agree will the sample be included in the final IntentQA dataset.  

\label{sec:Data Statistics}
\textbf{Dataset Statistics.} Upon completion of the filtering and annotation process, the main set of our IntentQA dataset comprises $4,303$ videos and $16,297$ question-answer (QA) pairs. This foundational dataset features $624$ distinct actions, $193$ lemmatized verbs, and $162$ action IDs, as summarized in \Cref{table:dataset_splitting}. We've structured the dataset into training, validation, and testing subsets following an approximate ratio of 6:1:1 to facilitate model training and evaluation. Specifically, the training set includes $12,119$ QAs, while the validation and testing sets contain $2,044$ and $2,134$ QAs, respectively, detailed in \Cref{table:dataset_splitting}.

In constructing the main set, we've taken careful measures to ensure the dataset's utility for training robust models. Each action's lemmatized verb that appears in either the validation or testing sets is guaranteed to appear at least twice in the training set, supporting effective learning and generalization. When an action's lemmatized verb is represented by a sufficient number of video samples, we adhere to the 6:1:1 distribution ratio across the three dataset splits as closely as possible. This approach minimizes the risk of overfitting by ensuring that videos are uniquely assigned to only one of the sets, thereby preventing any leakage of information between the training, validation, and testing phases.

\begin{table}[ht]
\centering 
\caption{Statistics of dataset splitting. \# VQA refers to the number of video question answering samples.} 
\resizebox{0.48\textwidth}{!}{%
\begin{tabular}{l|l||c|c|c|c}
\toprule
\multicolumn{2}{l||}{\textit{IntentQA}} & Training & Validation & Testing & Total\\
\midrule
\multicolumn{2}{l||}{\# Video} & 3212 & 524 & 567 & 4303\\
\multicolumn{2}{l||}{\# Action} & 605 & 399 & 397 & 624 \\
\multicolumn{2}{l||}{\# Lemmatized Verb} & 193 & 188 & 167 & 193 \\
\multicolumn{2}{l||}{\# Action ID} & 162 & 162 & 144 & 162 \\
\midrule
\multirow{4}{*}{\# VQA} & CW & 6989 & 1185 & 1250 & 9424 \\
                        & CH & 1940 & 334 & 359 & 2633 \\
                        & TP\&TN & 3190 & 525 & 525 & 4240 \\
                        & Total & 12119 & 2044 & 2134 & 16297 \\
\bottomrule
\end{tabular}%
}
\label{table:dataset_splitting}
\end{table}

\begin{figure*}[!t]
\begin{center}
   \includegraphics[width=0.95\textwidth]{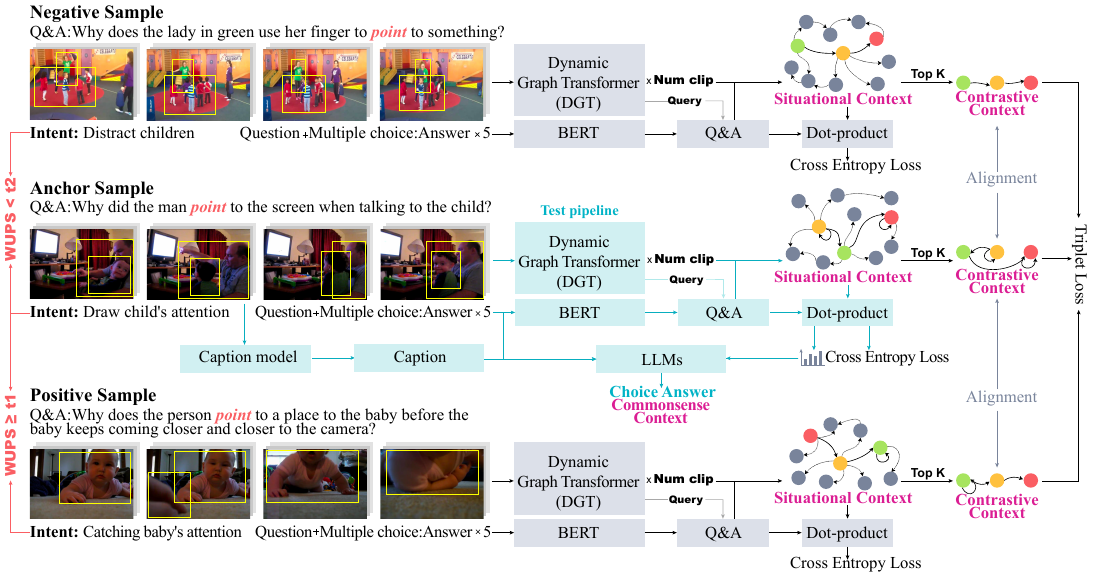}
\end{center}
   \caption{Overview of our eXplainable Context-aware Video Intent Reasoning (X-CaVIR). The figure contains a triplet of samples, i.e., the anchor sample, the positive sample and the negative sample. In the anchor sample, the agent guides the children to look at the screen. In the positive example, the agent guides the children to learn the skill of ``crawling''. In the negative example, the ``point'' action is a trick used by the agent in green to distract the children's attention and win the game. Our model utilizes situational context, contrastive context and commonsense context to solve the IntentQA task. The black color highlights the test pipeline. The yellow bounding boxes show the region features fed into our model.}
\label{fig:model}
\end{figure*}

\subsection{IntentQA Dataset Part II: the Contrast Set}
\begin{figure}[t]
    \centering
    \includegraphics[width=\linewidth]{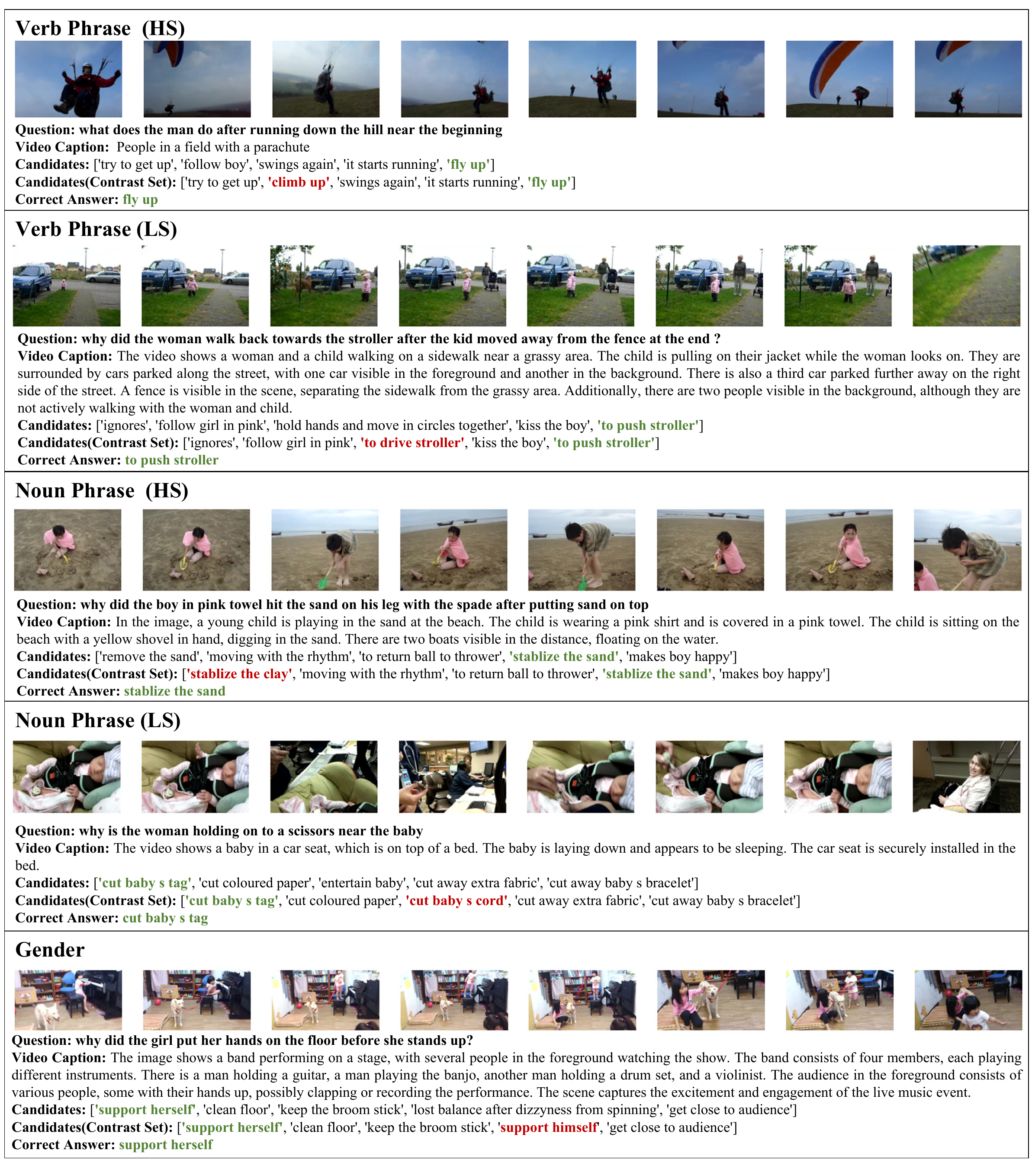} 

    \caption{\textcolor{black}{\textbf{Examples of contrast sets based on Verb Phrase, Noun Phrase, and Gender.} 
    To evaluate fine-grained reasoning, we generate contrastive distractors across three categories. 
    For Noun Phrases (NP) and Verb Phrases (VP), candidates are classified by semantic proximity:
    \textbf{High Similarity (HS)} denotes distractors that are semantically close to the correct answer (e.g., altering \textit{``stabilize the sand''} to \textit{``stabilize the clay''} [NP], or \textit{``fly up''} to \textit{``climb up''} [VP]), requiring precise visual discrimination.
    \textbf{Low Similarity (LS)} represents distinct actions or objects (e.g., \textit{``to push stroller''} vs. \textit{``to drive stroller''} [VP]).
    \textbf{Gender} contrast involves swapping pronouns or gendered nouns (e.g., \textit{``herself''} $\leftrightarrow$ \textit{``himself''}).}}
    \label{figure:contrast_set_diagram}
\end{figure}

Following the acquisition of the primary IntentQA dataset, we expanded it by constructing a contrast set comprising several distinct subsets. We adopted the methodology proposed by Park \textit{et al.}~\cite{park2022exposing}, which introduced a strategy for automatically generating text-based contrast sets for video-language tasks. \textcolor{black}{While their approach relied on traditional language models to perform relatively simple template-based substitutions, we significantly refined this methodology. We upgraded the pipeline to leverage the advanced capabilities of Large Language Models (LLMs), employing a more sophisticated and reliable process to ensure the high quality and effectiveness of the generated contrast options.} Furthermore, we extended the replacement criteria to encompass gender, verb phrases, and noun phrases. \textcolor{black}{Representative examples from these contrast subsets are visualized in \Cref{figure:contrast_set_diagram}, demonstrating the enhanced difficulty and robustness introduced by our modifications.}

\textbf{Gender Contrast Set}
The construction of the Gender Contrast Set begins by \textcolor{black}{referencing the gender-sensitive word mapping provided in the top section of \Cref{table:combined_contrast}.} For a given set of answers $\mathcal{A} = \{a_1, \ldots, a_i\}$, where $a^*$ denotes the correct answer, we \textcolor{black}{first verify whether $a^*$ contains any gender-sensitive terms listed in the table. If no such terms are present, the QA instance is excluded from this specific process. Conversely, if $a^*$ contains gender-sensitive words, we substitute each specific term with its counterpart of the opposite gender as defined in the mapping table.} This modification yields a gender contrast option, denoted as $a_c^g$.

\textcolor{black}{Subsequently, a non-correct answer $a^r$, randomly selected from the original set $\mathcal{A}$, is replaced by this generated option $a_c^g$. The updated answer set, $\mathcal{A}_c^g$, constitutes the final gender contrast set.} This methodology ensures that the dataset directly evaluates the QA system's sensitivity to gender-specific linguistic alterations.

\textbf{Verb Phrase and Noun Phrase Contrast Set}
For the generation of contrast sets involving verb and noun phrases, we maintain a consistent framework by selecting an incorrect answer $a^r$ from the original set $\mathcal{A}$ and replacing it with a generated contrast option. However, the procedure for creating these options \textcolor{black}{differs significantly from the deterministic gender approach. As shown in the bottom section of \Cref{table:combined_contrast}, we aim to generate distractors that vary in semantic similarity.}

Initially, we employ AllenNLP's dependency parsing~\cite{Dozat2017DeepBA} to analyze the correct answer $a^*$, identifying the \textcolor{black}{pivotal} verb or noun phrases. These phrases are then masked with a ``[mask]'' token. We subsequently prompt InstructGPT~\cite{ouyang2022training} to generate 20 potential replacement candidates for each mask, forming new options $a_c^v$ for verbs and $a_c^n$ for nouns. 

To ensure a comprehensive evaluation, \textcolor{black}{these generated options are categorized based on semantic similarity to the correct answer $a^*$. We select candidates from the top 15\% most semantically similar to form the \textit{High-Similarity (HS)} contrast set, and from the bottom 15\% to form the \textit{Low-Similarity (LS)} contrast set. As illustrated in \Cref{table:combined_contrast}, HS options (e.g., distinguishing ``put'' from ``perch'') pose a greater challenge by requiring fine-grained discrimination, whereas LS options (e.g., replacing ``keyboard'' with ``snack'') are more incongruent and easier to identify.} One option is randomly selected from each subset and added back into the original answer set $\mathcal{A}$, creating the respective high-similarity and low-similarity contrast sets.

\begin{table*}[!t]
    \centering
    \arrayrulecolor{black}
    \color{black}

    \caption{\textcolor{black}{\textbf{Data Construction Details.} Top: The mapping list used for gender-specific term conversion. Bottom: Some examples of contrast sets generated across different categories. \textbf{HS}: High Similarity, \textbf{LS}: Low Similarity. Changes are highlighted in \textcolor{red}{red}.}}
    \label{table:combined_contrast}

    \vspace{0.2cm}
    \resizebox{\textwidth}{!}{%
        \small
        \begin{tabular}{llll|llll}
            \toprule
            \multicolumn{8}{c}{\textbf{Gender-Specific Term Mappings}} \\
            \midrule
            he $\to$ she & his $\to$ her & man $\to$ woman & men $\to$ women & 
            she $\to$ he & her $\to$ his & woman $\to$ man & women $\to$ men \\
            boy $\to$ girl & boys $\to$ girls & guy $\to$ lady & guys $\to$ ladies & 
            girl $\to$ boy & girls $\to$ boys & lady $\to$ guy & ladies $\to$ guys \\
            \bottomrule
        \end{tabular}
    }

    \vspace{0.4cm} 

    \resizebox{\textwidth}{!}{%
        \small 
        \begin{tabular}{lccccc}
            \toprule
            \multirow{2}{*}{\textbf{Original}} & \multicolumn{2}{c}{\textbf{Verb Phrase (VP)}} & \multicolumn{2}{c}{\textbf{Noun Phrase (NP)}} & \multirow{2}{*}{\textbf{Gender}} \\
            \cmidrule(lr){2-3} \cmidrule(lr){4-5}
             & \textbf{High Similarity (HS)} & \textbf{Low Similarity (LS)} & \textbf{High Similarity (HS)} & \textbf{Low Similarity (LS)} & \\
            \midrule
            1. put keyboard on his lap & \textcolor{red}{perch} keyboard on his lap & \textcolor{red}{arrange} keyboard on his lap & put \textcolor{red}{book} on his lap & put \textcolor{red}{snack} on his lap & put keyboard on \textcolor{red}{her} lap \\
            \addlinespace[0.1cm]
            2. to face the camera & to \textcolor{red}{turn} the camera & to \textcolor{red}{open} the camera & to face the \textcolor{red}{curtain} & to face the \textcolor{red}{moon} & - \\
            \addlinespace[0.1cm]
            3. direct the plane & \textcolor{red}{choreograph} the plane & \textcolor{red}{guide} the plane & direct the \textcolor{red}{flight} & direct the \textcolor{red}{car} & - \\
            \addlinespace[0.1cm]
            4. hold them with his hands & \textcolor{red}{rock} them with his hands & \textcolor{red}{hug} them with his hands & hold them with his \textcolor{red}{embrace} & hold them with his \textcolor{red}{charm} & hold them with \textcolor{red}{her} hands \\
            \bottomrule
        \end{tabular}
    }

    \arrayrulecolor{black} 
    
\end{table*}

\section{Model}
\label{sec:model}
\subsection{Overview}
\label{sec: model overview}

We define the task of IntentQA to be the same as VideoQA in terms of input and output forms, taking a video $v$, a question $q$ and a corresponding answer set $\mathcal{A}$ as input, and outputting the correct answer $a^*$ from the answer set $\mathcal{A}$. 
\begin{equation}
a^* = \arg\max_{a\in \mathcal{A}}~f_{w}(a|q, v, \mathcal{A}),
\end{equation}
where $f_w$ represents a mapping function with learnable parameters $w$. Compared to traditional VideoQA tasks, the difference in our proposed intentQA task lies in that all the QAs are related to intent understanding. 

To solve this problem, we propose a eXplainable Context-aware Video Intent Reasoning (X-CaVIR), as shown in \Cref{fig:model}, which can sense context from three aspects. Firstly, we obtain the \textbf{Situational Context} from the video related to the question through VQL. Then, we select the positive and negative samples with the same action randomly, align the top-k highest attention nodes in the Situational Context, calculate the triplet loss, and obtain the \textbf{Contrastive Context}. Finally, we use GPT \cite{ouyang2022training} to obtain the \textbf{Commonsense Context}, and combine the predicted distribution of our model based on the Situational Context and Contrastive Context to get the final result. To further explain the overall structure of the model, we will start with a single sample.

For a single sample, we use a simplified version of VGT \cite{xiao2022video} as our baseline model. As shown in \Cref{fig:model_one_sample}, we use the frame features $V_f$ and the region features $V_r$ of the video as inputs. The region features $V_r$ are first modeled by $N$ DGTs \cite{xiao2022video} to form the region graph $G_r$:
\begin{equation}
G_{r} =  DGT(V_r),
\end{equation}
and then concatenated with $V_f$ to obtain the frame/region graph $G_{f,r}$:
\begin{equation}
G_{f,r} = {\rm Concat}(V_f, G_{r}),
\end{equation}
where DGT is from VGT \cite{xiao2022video}, and we use the same settings.
For the language part, we concatenated the questions and answers together and extract language features with Bert:
\begin{equation}
F_{q, \mathcal{A}} = {\rm Bert}({\rm Concat}(q, \mathcal{A})).
\end{equation}

Then, we use the region graph $G_r$ and the language feature $F_{q, \mathcal{A}}$, and employ the VQL model to extract the cross-modal graph $G_{r|q,\mathcal{A}}$, i.e., the \textbf{Situational Context} (see details in \Cref{sec:VQL}):
\begin{equation}
\label{eq: grqa}
G_{r|q,\mathcal{A}} = {\rm VQL}(G_r, F_{q, \mathcal{A}}).
\end{equation}

Next, we use the cross-modal graph $G_{r|q,\mathcal{A}}$ extracted by VQL to obtain \textbf{Contrastive Context} through contrastive learning.  
Finally, we use a multi-head self-attention (MHSA) transformer to fuse all the features and obtain a composite feature representation $F_{f,r|q,\mathcal{A}}$ of the video:
\begin{equation}
F_{f,r|q,\mathcal{A}} = {\rm MHSA}(G_{f,r}+G_{r|q,\mathcal{A}}).
\end{equation}

In \Cref{sec:Commonsense Reasoning}, we detail how we predict the results through \textbf{Commonsense Context} for the test pipeline.

\subsection{Video Query Language (VQL)}
\label{sec:VQL}
We use a video query language (VQL) approach to obtain visual contexts related to the question from the video. As shown in \Cref{fig:model_one_sample}, we use the video region graph $G_r$ obtained by extracting features from $N$ DGTs to query the QA features $F_{q, \mathcal{A}}$ extracted by BERT, and calculate the similarity matrix $S_{r|q, \mathcal{A}}$:
\begin{equation}
S_{r|q, \mathcal{A}} = G_rF_{q, \mathcal{A}}^ \intercal.
\end{equation}

Multiplying the similarity matrix $S_{r|q, \mathcal{A}}$ and the language feature $F_{q, \mathcal{A}}$, transforming the language feature $F_{q, \mathcal{A}}$ into the visual feature space $G_v$:
\begin{equation}
G_v = S_{r|q, \mathcal{A}} F_{q, \mathcal{A}},
\end{equation}
and then fusing them together to form the cross-modal graph $G_{r|q, \mathcal{A}}$, which is the question-relevant video contexts we need:
\begin{equation}
G_{r|q, \mathcal{A}} = G_r + G_v.
\end{equation}

\subsection{Contrastive Learning}
\label{sec:Contrastive Learning}

\begin{figure*}[!t]
\begin{center}
\includegraphics[width=0.85\textwidth]{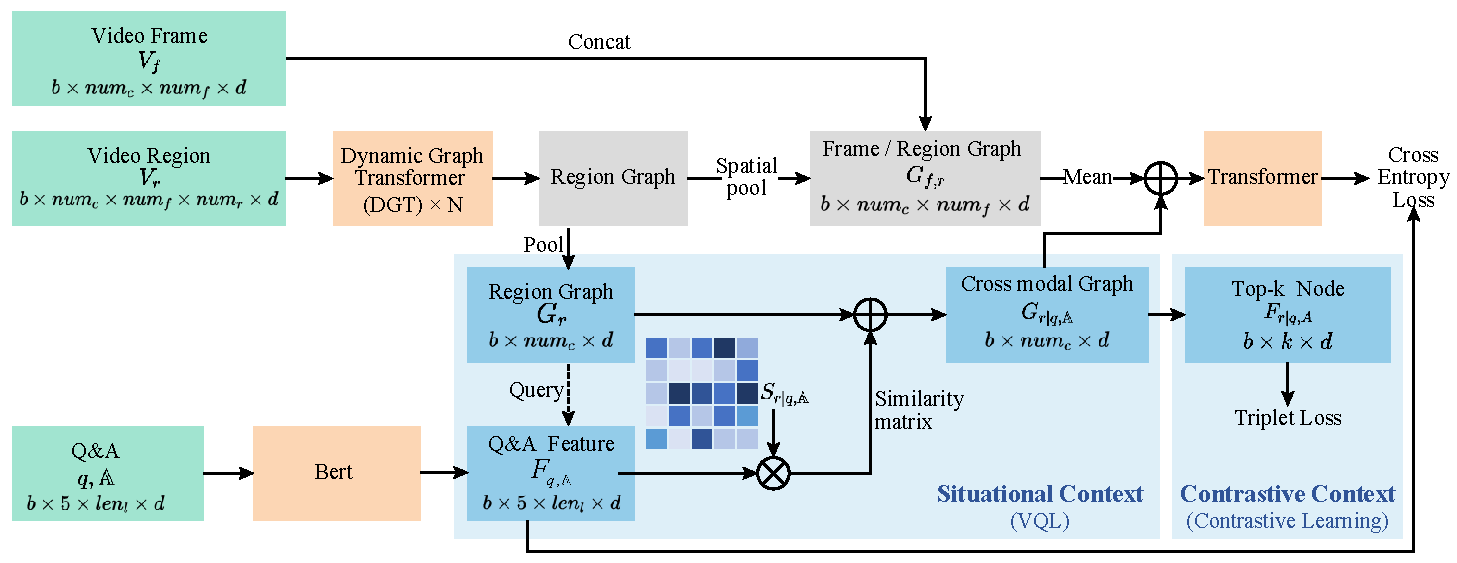}
\end{center}
   \vspace{-4pt}
   \caption{The model architecture for a single sample input. Green colors highlight the input. Orange highlights the modules borrowed from VGT. black highlights our new modules for extracting different contexts. $b$ represents the batch size. $num_c$ indicates the number of clips. $num_f$ denotes the number of frames in each clip. $num_r$ denotes the number of regions per frame. $d$ is the dimension of the features. $len_l$ is the length of Q\&A. $k$ refers to the top-k nodes selected.}
\label{fig:model_one_sample}
\end{figure*}

As shown in \Cref{fig:model}, we select positive and negative examples based on two similarity conditions between the action and answer of two QAs. To control the action similarity, we divide it into three levels according to action consistency/action's lemmatized verb consistency/action ID consistency. In order to determine whether two samples with the same action are positive or negative to each other, we compare the similarity of their answers via WUPS score \cite{malinowski2014multi}. As formula \Cref{eq:pos_neg} shows, when two QA samples, A and B, have a WUPS score between their correct answers ($a^*_A$,$a^*_B$) that is greater than or equal to a threshold  $t_1$, we consider A and B to be positive samples for each other; otherwise, if the WUPS score is below a threshold $t_2$, we regard A and B as negative samples of each other:
\begin{equation}
\label{eq:pos_neg}
\text{Relation}(A, B) = 
\begin{cases}
\text{Pos.} & \text{WUPS ($a^*_A$,$a^*_B$) $\geq t_1$},\\
\text{Neg.} & \text{WUPS ($a^*_A$,$a^*_B$) $< t_2$}.
\end{cases}
\end{equation} 

We collect positive and negative examples for each QA to allow the anchor sample to randomly select one positive and one negative example to form a triplet as the input.

As illustrated in \Cref{fig:model_one_sample}, we initially extract the features $F_{r|q, \mathcal{A}}$ corresponding to the top-k nodes from the cross-modal graph $G_{r|q, \mathcal{A}}$. These nodes are selected based on their highest relevance to the given question and answer set, as determined by the similarity matrix $S_{r|q, \mathcal{A}}$:
\begin{equation}
F_{r|q, \mathcal{A}} = {\rm top\text{-}k}(G_{r|q, \mathcal{A}}, S_{r|q, \mathcal{A}}).
\end{equation}

We repeat this operation for the three samples in the triplet to obtain $F_{r|q,\mathcal{A}}^{a}$, $F_{r|q,\mathcal{A}}^{p}$, $F_{r|q,\mathcal{A}}^{n}$.

Then we align the features of the negative example $F_{r|q, \mathcal{A}}^{n}$ and the positive example $F_{r|q, \mathcal{A}}^{p}$ to the anchor sample:
\begin{equation}
\begin{aligned}
{F_{r|q, \mathcal{A}}^{p}}^{\text{align}} = F_{r|q, \mathcal{A}}^{a}{F_{r|q, \mathcal{A}}^{p}}^{\intercal}F_{r|q, \mathcal{A}}^{p}, \\
{F_{r|q, \mathcal{A}}^{n}}^{\text{align}} = F_{r|q, \mathcal{A}}^{a}{F_{r|q, \mathcal{A}}^{n}} ^{\intercal}F_{r|q, \mathcal{A}}^{n}.
\end{aligned}
\end{equation}

The distance between the anchor sample and the positive/negative samples, $d(a, p)$ and $d(a, n)$ are computed as:
\begin{equation}
\begin{aligned}
d(a, p) = (F_{r|q, \mathcal{A}}^{a} - {F_{r|q, \mathcal{A}}^{p}}^{\text{align}})^2, \\
d(a, n) = (F_{r|q, \mathcal{A}}^{a} - {F_{r|q, \mathcal{A}}^{n}}^{\text{align}})^2. 
\end{aligned}
\end{equation}

The triplet loss is:
\begin{equation}
\label{eq:triplet loss}
L_\text{triplet} = \max(d(a, p) - d(a, n) + \text{margin},0).
\end{equation}

For each sample of the triplet, the language feature $F_{q, \mathcal{A}}$ and the composite feature $F_{f,r|q, \mathcal{A}}$ after the global MHSA transformer are calculated by dot product to obtain the matching scores $S$ of the answer set $\mathcal{A}$:
\begin{equation}
S = F_{f,r|q, \mathcal{A}}F_{q, \mathcal{A}}^\intercal.
\label{eq:matching score}
\end{equation}

Then we calculate the cross-entropy loss:
\begin{equation}
L_\text{ce} = -\sum_{i=1}^{|\mathcal{A}|}y_{i}\log S_{i}.
\end{equation}

The matching score $S$ is calculated as \Cref{eq:matching score}. \( y_i = 1 \) if the index of the predicted answer matches the ground-truth answer for the \( i \)-th sample; otherwise, \( y_i = 0 \). The complete loss $L$ calculated as:
\begin{equation}
L = L_\text{ce}^a + L_\text{ce}^p + L_\text{ce}^n + L_\text{triplet}.
\end{equation}

\subsection{Commonsense Reasoning}
\label{sec:Commonsense Reasoning}

We propose a method that effectively leverages the commonsense reasoning capabilities of Large Language Models (LLMs) like GPT \cite{ouyang2022training} during the inference stage. \textcolor{black}{In our preliminary conference version \cite{Li2023IntentQA}, we employed a \textit{Late Fusion} strategy, which simply performed a linear combination of the prediction distribution from a blind LLM (text-only) and the matching scores from the visual model. While effective, that approach treated the LLM merely as a separate scoring branch, failing to fully utilize its ability to reason over visual uncertainty. To address this, we introduce a \textit{Confidence Prompting} pipeline in this work. Instead of merging scores post-hoc, we explicitly embed the visual model's matching scores into the prompt, allowing the LLM to act as a reasoner that considers both its internal commonsense and the visual model's confidence.}

To implement \textit{Confidence Prompting}, we first format the visual model's output into a natural language representation. We concatenate the answer set $\mathcal{A}$ and the matching scores $S$ obtained from \Cref{eq:matching score} to generate the candidate options $C$ annotated with confidence levels:
\begin{equation}
C = {\rm Concat}(a_i, S_i \mid a_i \in \mathcal{A}, S_i \in S).
\end{equation}

This step corresponds to the input for Model 6 in our ablation studies. To further enhance the context for Model 7 (our full model), we generate a textual description of the visual content. We input the video frames $V_f$ into a caption model to obtain the video description $D$:
\begin{equation}
D = f_{\text{caption}}(V_f).
\end{equation}

In the setting of this paper, the caption model we use is Qwen 7B \cite{Qwen-VL}.

Finally, we construct a comprehensive prompt containing the confidence-embedded candidates $C$, the video description $D$, and the question $Q$. This is fed into the LLM to select the most likely answer and provide a reasoning process $R$:
\begin{equation}
a^*, R = LLM(C, D, Q).
\label{eq:common sense reasoning}
\end{equation}

As shown in \Cref{fig:intro} (b), we illustrate this pipeline in detail. First, we use our Context-Aware Video QA model to obtain the matching score $S$ of each option. These are then recombined with the option text to form ``Candidates with Confidence'' $C$, formatted as ``[$C_0(S_0), C_1(S_1), \ldots$]''. Simultaneously, for the full model setup, we employ a caption model to obtain dense captions $D$ detailing the content of each frame sequence.

Then, we input these components into the LLM, requesting a response in a specified format. The response requirements include three items: ``Option ID,'' ``Option,'' and ``Reasoning Process.'' It is worth mentioning that, when evaluating the model's performance, we consider a response legitimate only if the ``Option ID'' corresponds accurately to the ``Option''; otherwise, we deem the response incorrect. The reason for additionally requiring ``Option'' is to avoid overestimating the LLM's performance in cases where it might simply guess the correct answer by chance. Additionally, we require the LLM to provide a complete ``Reasoning Process,'' which not only allows for improved performance through the use of Chain of Thought (CoT) reasoning but also offers a chance for us to qualitatively assess the LLM's reasoning ability. The ``Reasoning Process'' provided by the LLM allows us to easily identify the reasons for failure in failed cases, providing interpretability to the model's reasoning in linguistic form and offering insights for subsequent model optimization.

\section{Experiments}\label{sec:experiment}
\subsection{Ablation Experiments}
\subsubsection{Model Component Diagnosis}
\label{sec:individual components}

\textcolor{black}{To facilitate a clear analysis of our architectural evolution and the impact of different LLM backbones, we index our models as follows: Integer IDs (e.g., 5, 6, 7) denote distinct fusion pipelines or component combinations. Specifically, Model 5 represents the baseline \textit{Late Fusion} strategy from our conference version \cite{Li2023IntentQA}. Model 6 denotes our proposed \textit{Confidence Prompting} pipeline without visual captions, whereas Model 7 represents the full model incorporating caption information. Suffixes (e.g., -1, -2) indicate the substitution of the LLM backbone with ChatGPT and GPT-4, respectively.}

To assess the effectiveness of our essential components, we design the following comprehensive ablation experiments, as shown in \Cref{tab:Ablation Experiments.}. \textbf{``Blind LLM''} only uses GPT \cite{ouyang2022training} for the IntentQA task, and thus with no video input. \textbf{``Base Model''} is a simplified VGT model. \textbf{``+ VQL''} adds Video Query Language onto the base model to get a cross-modal graph for better situational context representation. \textbf{``+ Triplet Loader''} loads the anchor sample together with the positive and negative samples during training. \textbf{``+ Triplet Loss''} continues to add triplet margin loss. \textbf{``+ LLM''} adds commonsense prior of LLM during the test. The \textbf{``+ Caption''} component introduces additional information. It further offers more reasoning evidence from the video description, facilitating the LLM in establishing more logical reasoning chains. The visual backbone components (Models 1-4) are \textbf{subsequently} and \textbf{cumulatively} added. Regarding LLM integration, Models 5 and 6 represent distinct strategies built upon the visual backbone: Model 5 applies the baseline Late Fusion, while Model 6 introduces our proposed Confidence Prompting. Finally, Model 7 builds upon Model 6 by incorporating captions.

As shown in \Cref{tab:Ablation Experiments.}, our entire model achieved the best performance across all tests, with each component significantly contributing to the performance improvement. Integrating the commonsense capabilities of LLMs into our model resulted in the most significant boost, bringing a $+3.98\%$ increase. \textcolor{black}{Furthermore, we examine the effectiveness of our proposed pipeline by isolating the fusion strategy. Comparing Model 5 and Model 6—both utilizing the same InstructGPT backbone without dense captions—reveals that our Confidence Prompting method (Model 6, 58.48\%) outperforms the Late Fusion strategy employed in our conference version (Model 5, 57.64\%). This demonstrates that the performance gain stems from the enhanced reasoning structure rather than merely the choice of LLM or auxiliary inputs.} Further incorporation of a caption model, enabling the LLM to gather more evidence for constructing reasoning chains, leads to the second-largest gain of $+2.25\%$. Fundamental modifications to the base VQA model also contributed to performance improvements, as demonstrated by the first four ablation studies. Additionally, among all the experiments, \textit{TP\&TN}-type QAs proved to be the most challenging compared to \textit{CW} and \textit{CH}, possibly because intents are not explicitly expressed in \textit{TP\&TN}-type QAs.

\begin{table*}[t!bp]
\begin{center}
\small{
\resizebox{0.85\textwidth}{!}{%
\begin{tabular}{clcccccccccccc}
\toprule
 \multirow{2}{*}{Model ID}& \multirow{2}{*}{Model} & \multicolumn{2}{c}{\textit{CW}} & &\multicolumn{2}{c}{\textit{CH}}& &\multicolumn{2}{c}{\textit{TP\&TN}} & &\multicolumn{2}{c}{Total}\\
\cline{3-4} \cline{6-7} \cline{9-10} \cline{12-13}
  & & Val. & Test &  & Val. & Test &  & Val. & Test &  & Val. & Test \\
\midrule
\midrule
0 & Blind LLM(InstructGPT) & - & 52.16 & & - & 61.28 &  & - & 43.43 &  &- &51.55\\
\midrule
1 & Base Model & 50.89 & 51.76 & & 54.79 & 56.27 &  & 48.00 & 47.05 &  &50.78 &51.36\\
2 & + VQL & 51.65 & 52.32 & & 54.49 & 58.77 &  & 47.62 & 48.00 &  &51.08 &52.34 (+0.98)\\
3 & + Triplet Loader & 51.56 & 53.60 & & 56.89 & 60.72 &  & 48.00 & 49.52 &  &51.52 &53.80 (+1.46)\\
4 & + Triplet Loss & 52.57 & 55.28 & & 57.47 & 61.56 &  & 46.10 & 47.81 &  &51.71 &54.50 (+0.70)\\
\midrule
\textcolor{black}{5} &\textcolor{black}{\textbf{+ InstructGPT (Late Fusion) \cite{Li2023IntentQA}}} & \textcolor{black}{-} & \textcolor{black}{\textbf{58.40}} & & \textcolor{black}{-} & \textcolor{black}{\textbf{65.46}} &  & \textcolor{black}{-} & \textcolor{black}{\textbf{50.48}} &  &\textcolor{black}{-} &\textcolor{black}{\textbf{57.64 (+3.14)}} \\
\textcolor{black}{6} &\textcolor{black}{\textbf{+ InstructGPT (Conf. Prompt)}} & \textcolor{black}{-} & \textcolor{black}{\textbf{58.48}} & & \textcolor{black}{-} & \textcolor{black}{\textbf{66.02}} &  & \textcolor{black}{-} & \textcolor{black}{\textbf{53.33}} &  &\textcolor{black}{-} &\textcolor{black}{\textbf{58.48 (+3.98)}} \\
\textcolor{black}{7} &\textcolor{black}{\textbf{+ Caption (Full Model)}} & \textcolor{black}{-} & \textcolor{black}{\textbf{60.95}} & & \textcolor{black}{-} & \textcolor{black}{\textbf{67.69}} &  & \textcolor{black}{-} & \textcolor{black}{\textbf{55.43}} &  &\textcolor{black}{-} &\textcolor{black}{\textbf{60.73 (+2.25)}} \\
\bottomrule
\end{tabular}}}
\end{center}
\vspace{-0.4cm}
\caption{Ablation diagnosis of our model components. We use accuracy (\%) as the metric. ``+'' indicates adding new components on top of the previous row. \textcolor{black}{Model 5 serves as the baseline from our conference version. Models 6 and 7 employ our proposed Confidence Prompting pipeline, with Model 7 further integrating caption information.}}
\label{tab:Ablation Experiments.}
\vspace{-0.4cm}
\end{table*}

\begin{table*}
\begin{center}
\scriptsize{
\resizebox{0.7\textwidth}{!}{%
\begin{tabular}{clcccccccccccc}
\toprule
  \multirow{2}{*}{Model ID}& \multirow{2}{*}{Model} & \multicolumn{2}{c}{\textit{CW}} & &\multicolumn{2}{c}{\textit{CH}}& &\multicolumn{2}{c}{\textit{TP\&TN}} & &\multicolumn{2}{c}{Total}\\
\cline{3-4} \cline{6-7} \cline{9-10} \cline{12-13}
  & & Val. & Test &  & Val. & Test &  & Val. & Test &  & Val. & Test \\
\midrule
\midrule
2 & Base Model + VQL & 51.65 & 52.32 & & 54.49 & 58.77 &  & 47.62 & 48.00 &  &51.08 &52.34\\
\hline
4-1 & Action ID,  $t1=0.85$  & 51.48 & 52.32 & & 50.60 & 58.77 &  & 48.38 & 47.62 &  &50.54 &52.25\\
4-2 & Action ID, $t1=1$  & 50.13 & 53.12 & & 54.49 & 57.66 &  & 47.24 & 48.19 &  &50.10 &52.67\\
4-3 & Lemmatized Verb, $t1=0.85$   & 51.90 & 53.36 & & 55.99 & 62.12 &  & 44.00 & 45.52 &  &50.54 &52.91\\
4-4 & Lemmatized Verb, $t1=1$  & 50.80 & 54.56 & & 55.09 & \textbf{62.67} &  & 48.00 & 46.10 &  &50.78 &53.84\\
4-5 & Action, $t1=0.85$ & 50.21 & 52.00 & & 56.29 & 59.61 &  & \textbf{48.95} & \textbf{49.52} &  &50.88 &52.67\\
4 & Action, $t1=1$ & \textbf{52.57} & \textbf{55.28} & & \textbf{57.47} & 61.56 &  & 46.10 & 47.81 &  &51.71 &54.50\\
\midrule
4-6 & Mask Randomly & 50.89 & 51.28 & & 53.89 & 56.27 &  & 45.14 & \underline{48.76} &  &49.90 &51.50\\
4-7 & Mask Lowest Attention & \underline{52.83} & \underline{54.72} & & \underline{57.49} & \underline{59.89} &  & \underline{46.86} & 48.38 &  &\underline{52.05} & \underline{54.05}\\
\bottomrule
\end{tabular}}}
\end{center}
\vspace{-0.4cm}
\caption{Analysis of contrastive learning (best shown in bold) and context attention (best shown with underline). We use accuracy (\%) as the metric. }
\label{tab:contrastive learning analysis}
\vspace{-0.4cm}

\end{table*}
\begin{table*}[ht]
\begin{center}
\scriptsize{
\resizebox{0.7\textwidth}{!}{%
\begin{tabular}{cllcccccccc}
\toprule
  \multirow{1}{*}{Model ID}& \multirow{1}{*}{Model} & & \multicolumn{1}{c}{\textit{CW}} & &\multicolumn{1}{c}{\textit{CH}}&&\multicolumn{1}{c}{\textit{TP\&TN}} & &\multicolumn{1}{c}{Total}\\
\midrule
\midrule
\textcolor{black}{7} & \textcolor{black}{Ours full (InstructGPT)}  && 60.95 & & 67.69 & & 55.43 & & 60.73 &  \\
\textcolor{black}{7-1} & \textcolor{black}{Ours full (ChatGPT)}  && 59.28 & & 67.97 & & 53.33 & & 59.28 & \\
\textcolor{black}{7-2} & \textcolor{black}{Ours full (GPT-4)} & & 69.90 & & 68.80 & & 50.67 & & \textbf{64.81} &  \\
\textcolor{black}{6} & \textcolor{black}{Ours (InstructGPT) w/o Caption} & & 58.48 & & 66.02 & & 53.33 & & 58.48(-2.25) &  \\
\textcolor{black}{6-1} & \textcolor{black}{Ours (ChatGPT) w/o Caption} & & 57.92 & & 65.46 & & 50.86 & & 57.45(-1.83) &  \\
\textcolor{black}{6-2} & \textcolor{black}{Ours (GPT-4) w/o Caption} && 63.28 & & 72.98 & & 54.10 & & 62.65(-2.16) &  \\
8 & Caption + InstructGPT && 56.56 & & 58.22 & & 45.14 & & 54.03(-6.7) &  \\
8-1 & Caption + ChatGPT& & 57.36 & & 64.35 & & 44.95 & & 55.48(-3.8)&  \\
8-2 & Caption + GPT-4 & & 68.16 & & 67.13 & & 48.76 & & 63.21(-1.6) &  \\
\bottomrule
\end{tabular}}}
\end{center}
\vspace{-0.4cm}
\caption{Ablation analysis on integrating LLM for common sense reasoning. We use accuracy (\%) as the evaluation metric. ``-'' indicates removing certain components from \textcolor{black}{Model 7 (Ours full)}. }
\label{tab:commonsense reasoning analysis}
\vspace{-0.5cm}
\end{table*}

\begin{table*}[ht]
\begin{center}
\resizebox{\textwidth}{!}{
\begin{tabular}{cllc ccccccccccccccc}
\toprule
  \multirow{2}{*}{Model ID}& \multirow{2}{*}{Model} &\multirow{2}{*}{Text Rep.} & \multirow{2}{*}{LLM Integ.} & \multicolumn{2}{c}{\textit{CW}} & &\multicolumn{2}{c}{\textit{CH}}& &\multicolumn{2}{c}{\textit{TP\&TN}} & &\multicolumn{2}{c}{Total}\\
\cline{5-6} \cline{8-9} \cline{11-12} \cline{14-15}
  & & & & Val. & Test &  & Val. & Test &  & Val. & Test &  & Val. & Test \\
\midrule
\midrule
- & EVQA \cite{antol2015vqa}  &GloVe & - & 25.99 & 25.92 & & 37.43 & 34.54 &  & 28.00 & 25.52 &  &28.38 &27.27\\
- & CoMem \cite{gao2018motion}  &GloVe & - & 31.56 & 30.00 & & 35.63 & 28.69 &  & 28.57 & 28.95 &  &31.46 &29.52\\
- & HGA \cite{jiang2020reasoning} &GloVe & - & 29.45 & 32.00 & & 35.03 & 30.64 &  & 29.71 & 31.05 &  &30.43 &31.54\\
- & HME \cite{fan2019heterogeneous} &GloVe & - & 30.97 & 34.40 & & 35.33 & 34.26 &  & 34.29 & 29.14 &  &32.53 &33.08\\
- & HQGA \cite{xiao2022HQGA} &GloVe & - & 32.49 & 33.20 & & 38.32 & 34.26 &  & 34.48 & 36.57 &  &33.95 &34.21\\
- & CoMem \cite{gao2018motion} &BERT & - & 46.75 & 47.68 & & 57.49 & 54.87 &  & 41.71 & 39.05 &  &47.21 &46.77\\
- & HGA \cite{jiang2020reasoning} &BERT & - & 43.54 & 44.88 & & 56.89 & 50.97 &  & 42.48 & 39.62 &  &45.45 &44.61\\
- & HME \cite{fan2019heterogeneous} &BERT & - & 46.50 & 46.08 & & 51.20 & 54.32 &  & 44.76 & 40.76 &  &46.82 &46.16\\
- & HQGA \cite{xiao2022HQGA} &BERT & - & 45.91 & 48.24 & & 57.78 & 54.32 &  & 44.76 & 41.71 &  &47.55 &47.66\\
- & VGT \cite{xiao2022video} &BERT & - & 50.46 & 51.44 & & 55.99 & 55.99 &  & 48.19 & 47.62 &  &50.78 &51.27\\
- & Blind LLM(InstructGPT) \cite{ouyang2022training} & - & - & - & 52.16 & & - & 61.28 &  & - & 43.43 &  &- &51.55\\
\midrule
\textcolor{black}{5} & \textcolor{black}{Ours w/o Caption (InstructGPT) \cite{Li2023IntentQA}} &\textcolor{black}{BERT} & \textcolor{black}{Late Fusion} & \textcolor{black}{-} & \textcolor{black}{58.40} & & \textcolor{black}{-} & \textcolor{black}{65.46} &  & \textcolor{black}{-} & \textcolor{black}{50.48} &  &\textcolor{black}{-} &\textcolor{black}{57.64}\\
\textcolor{black}{6} & \textcolor{black}{Ours w/o Caption (InstructGPT)}  &\textcolor{black}{BERT} & \textcolor{black}{Conf. Prompt} & - &58.48 & & - & 66.02 &  & - & 53.33&  &- &58.48\\
\textcolor{black}{7} & \textcolor{black}{Ours (InstructGPT)}  &\textcolor{black}{BERT} & \textcolor{black}{Conf. Prompt} & - & 60.95 & & - & 67.69 &  & - & 55.43 &  &- &60.73\\
\textcolor{black}{6-2} & \textcolor{black}{Ours w/o Caption (GPT4)}  &\textcolor{black}{BERT} & \textcolor{black}{Conf. Prompt} & - & 63.28 & & - & 72.98 &  & - & 54.10 &  &- &62.65\\
\textcolor{black}{7-2} & \textcolor{black}{\textbf{Ours (GPT4)}}  &\textcolor{black}{BERT} & \textcolor{black}{Conf. Prompt} & - & \textbf{69.90} & & - & \textbf{68.80} &  & - & \textbf{50.67} &  &- &\textbf{64.81}\\
\rowcolor{gray!10}
- & Human  &- & - &- &77.76  & &-  &80.22  &  &-  &79.05  &  &- &78.49\\
\bottomrule
\end{tabular}}
\end{center}
\caption{Comparison results with the established VideoQA baseline models. We use accuracy (\%) as the metric. \textcolor{black}{The ``LLM Integ.'' column denotes the method used to combine visual information with the LLM. Model 5 corresponds to our conference version \cite{Li2023IntentQA} using Late Fusion, while our current models employ Confidence Prompting. For detailed configurations, please refer to Table \ref{tab:Ablation Experiments.}.}}
\label{tab:Compare with Established}
\end{table*}

\begin{table}
\begin{center}
\scriptsize{
\resizebox{0.46\textwidth}{!}{
\begin{tabular}{clcc}
\toprule
   \multirow{2}{*}{Model ID}& \multirow{2}{*}{Model} & \multicolumn{2}{c}{MSRVTT-QA (OE)} \\
\cline{3-4}
  & & Val. & Test  \\
\midrule
\midrule
- & VGT \cite{xiao2022video} & 38.26 & 39.00 \\
3 & Ours (w/o triplet loss) & 38.11 & 39.21\\
4 & Ours (w/ triplet loss) & \textbf{38.98} & \textbf{39.39}\\
\bottomrule
\end{tabular}}}
\end{center}
\caption{Generalization test on dataset \textit{Open-ended MSRVTT-QA}. We use accuracy (\%) as the metric.}
\label{tab:finetuning}
\end{table}

\subsubsection{Contrastive Learning Analysis}
\label{sec:contrastive Learning Analysis}
To further verify our contrastive learning approach, we utilize Model 4, which is the best-performing model without LLM integration, to analyze how the selection criterion for contrastive samples would influence the performance. We control two factors for selecting positive and negative samples: (1) What is used to calculate the action similarity, which could be ``Action'', ``Lemmatized Verb'' or ``Action ID''; (2) The WUPS score threshold for answer similarity. We set the threshold $t_2$ in \Cref{eq:pos_neg} to 0.5, and discuss on the value of threshold $t_1$, i.e., $t_1=0.85$ or $t_1=1$. As shown in \Cref{tab:contrastive learning analysis}, model 4 (``Action, $t1=1$'') achieves the best performance. The result indicates the stricter criterion of action/answer similarity, the better performance.

\subsubsection{Context Attention Analysis}
\label{sec:context attention}

In order to verify whether our model learns to extract the most significant context information to solve the IntentQA task, we add two more analysis experiments: (i) \textbf{Mask Randomly}. We randomly mask $k$ nodes of the cross-modal graph ($G_{r|q,\mathcal{A}}$, see \Cref{eq: grqa}). (ii) \textbf{Mask Lowest Attention}. We mask the bottom $k$ nodes of the cross-modal graph with the lowest attention. As shown in \Cref{tab:contrastive learning analysis}, randomly masking the nodes severely hurts the model performance (decrease from $54.5$ to $51.5$), while masking the nodes with the lowest attention only influences the model performance very slightly (decrease from $54.5$ to $54.05$). The results verify that our model's capability in paying attention to the most valuable parts of the context.

\subsubsection{Commonsense Reasoning Analysis}
\label{sec:Commonsense Reasoning Analysis}
To investigate the impact of different information channels on model performance in Commonsense Reasoning, we carried out additional ablation experiments by removing either the caption (\textcolor{black}{Model 6}) or the VideoQA likelihood (Model 8) components from the full model (\textcolor{black}{Model 7}). As demonstrated in \Cref{tab:commonsense reasoning analysis}, eliminating either the VideoQA likelihood or captions led to a performance decline compared to the complete model, with the omission of the VideoQA likelihood resulting in a more pronounced decrease. This reveals that both the VideoQA likelihood and captions enhance the overall model's performance, with a more significant influence attributed to the VideoQA likelihood.

Moreover, we assessed the impact of substituting different LLMs under identical conditions. The outcomes, illustrated in \Cref{tab:commonsense reasoning analysis}, indicate that employing ChatGPT and InstructGPT yielded comparable enhancements in performance, while the use of GPT-4 brought about a substantial improvement. Specifically, GPT-4 achieved the highest overall performance with an accuracy of 64.81\%, followed by InstructGPT at 60.73\%, and ChatGPT at 59.28\%. This highlights the pivotal role of an LLM's reasoning capabilities in the overall method's efficacy, with GPT-4 outperforming ChatGPT and InstructGPT in harnessing commonsense reasoning.

Further analysis shows that the VideoQA likelihood has a significant impact on the performance of InstructGPT and ChatGPT, with performance declines of 6.7\% and 3.8\% respectively when this component is removed. In contrast, GPT-4 exhibited a smaller performance drop of 1.6\%, indicating its greater resilience to the absence of this information channel. In terms of the impact of captions, the performance decline was relatively consistent across different LLMs, with decreases of 2.25\% for InstructGPT, 1.83\% for ChatGPT, and 2.16\% for GPT-4. This suggests that while captions are important, their influence is less variable across different LLMs compared to the VideoQA likelihood.

The model's increased reliance on its reasoning capabilities over the VideoQA likelihood's suggestions, especially as these capabilities are enhanced, is evidenced by the last three rows of \Cref{tab:commonsense reasoning analysis}. This trend suggests a diminishing impact of removing the VideoQA likelihood on the model's performance as the LLM's reasoning strength intensifies. Similar findings were observed in our contrast set experiments, which are elaborated upon in \Cref{tab:Ablation Contrast}.

\subsection{Comparison with VideoQA Baselines}
\label{sec:videoqa baselines}

We compare our full model with several established VideoQA baseline models, as shown in \Cref{tab:Compare with Established}. We select several established VideoQA models from 2015 to 2022 as the baselines, including \textit{EVQA} \cite{antol2015vqa} proposed for the earliest VQA task, \textit{CoMem} \cite{gao2018motion} and \textit{HME} \cite{fan2019heterogeneous} using memory modules to model visual appearance, motion and language, as well as \textit{HGA} \cite{jiang2020reasoning}, \textit{VGT} \cite{xiao2022video} and \textit{HQGA} \cite{xiao2022HQGA}  using the graph to model videos. These selected baseline models respectively represent several typical methods for VideoQA.

As shown in \Cref{tab:Compare with Established}, our full model performs the best, and our model without GPT performs the second best. The early VideoQA models may focus on QA about video content description, i.e., the factoid VideoQA, they perform poorly on our IntentQA task, which requires better reasoning abilities of the unobservable intent. However, even the most recent SOTA models \textit{VGT} and \textit{HQGA} still have a large performance gap with our model. Contrastive situational context effectively improves model performance on \textit{CW} and \textit{CH} QA, but only slightly improves the performance on \textit{TP\&TN} QA. Commonsense context further significantly improve the model performance in all types of QA tasks.

In addition, we reported human results in \Cref{tab:Compare with Established}\footnote{We performed our human study with IRB approval.}, which are far superior to our model and other established models. This indicates that compared to existing models, humans still have a great advantage in understanding the intentions of humans in social contexts. At the same time, this also highlights the importance of the task we proposed, and the exploration of model understanding of social intentions and human cognition is still in its early stages. This problem is distinct from the traditional video understanding problem. It comprehends the video from the perspective of human cognition, exploring the hidden human intentions beneath the surface visual context, providing a novel perspective for video understanding.

\begin{figure*}[tbp]
\begin{center}
   \includegraphics[width=1\textwidth]{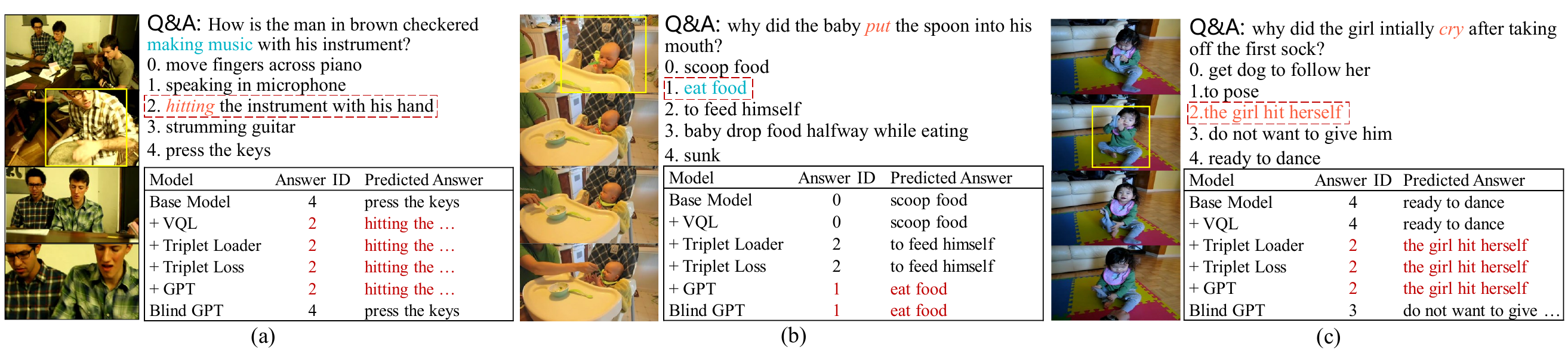}
\end{center}
   \caption{Qualitative Results and Analysis. The yellow boxes highlight the evidence contexts used to determine the correct answer. The red box frames the correct answer. Actions are colored in red while intents are colored in black.}
\label{fig:visualize}
\end{figure*}

\subsection{Generalization Test on Traditional VideoQA Task}
\label{sec:generalization}
We test our IntentQA model's generalization ability to other VideoQA tasks. We choose a large-scale open-ended VideoQA dataset \textit{MSRVTT-QA}, which contains 244k descriptive QA pairs and is a challenging traditional factoid VideoQA dataset, different from our inference VideoQA dataset. All the models, i.e., \textit{VGT}, \textit{Ours (w/o triplet loss)} and \textit{Ours (w/ triplet loss)}, are pre-trained on our \textit{IntentQA} dataset, and then finetuned on \textit{MSRVTT-QA}. \Cref{tab:finetuning} shows the results. Both of our two models achieve better accuracy than the baseline, and the model with triplet loss generalizes better. The test verifies our conjecture that intent reasoning and understanding based on contrastive situational context would help the model to better understand the video contexts, and generalize well to a new factoid VideoQA task.  

\subsection{Qualitative Results and Analysis}
\label{sec:Qualitative Analysis}

\textbf{How Does VQL Work?}
In the example illustrated in \Cref{fig:visualize} (a), there are three men playing different instruments. To answer the question correctly, the model needs to understand that the question is asking about `the man in brown checkered', and pay attention to the correct context in the video while ignoring other contexts. Our base model gets the wrong answer, but our model with VQL successfully predicts the correct answer. \textit{Blind LLM} could not answer correctly without any video context input.

\textbf{How Does \textbf{Commonsense Context} Work?}
In the example shown in \Cref{fig:visualize} (b), the question asks why the child put the spoon into his mouth, and the candidates ``scoop food'', ``eat'', ``feed'', and ``drop food'' all appear in the video, which might confuse the model a lot. The basic two models simply choose the most obvious action ``scoop food'' in the video. The two models with contrastive learning correctly understand that the subject is the baby, but still get the wrong answer. Note that it's the mother that is feeding the baby, thus the most appropriate answer is ``eat food''. The slight difference between ``feed'' and ``eat'' requires deep commonsense knowledge, and thus only our model with GPT and \textit{Blind LLM} get the answer right. \textit{Blind LLM} can even answer correctly based solely on text and commonsense without the video context, just as humans do.

\textbf{How Does Contrastive Learning Work?}
As shown in \Cref{fig:visualize} (c), to answer the question ``why the girl cry after taking off the first sock'', \textit{Blind LLM} guessed the answer to be `do not want to give him' based on commonsense, but it's wrong. The two basic models with situational context exaggerate the girl's physical movement and choose the wrong answer ``ready to dance''. The real context causing ``cry'' is very subtle, being the instant moment when the girl hit foot on ground after taking off the sock. Through contrastive learning with other positive and negative samples, the model learns that usually cry is caused by injury; thus the three models with contrastive context are correct.

\begin{table*}[!tbp]
\begin{center}
\scriptsize{
\resizebox{\textwidth}{!}{%
\begin{tabular}{cllccccc}
\toprule
 \multirow{2}{*}{ID} & \multirow{2}{*}{Model} & \multirow{2}{*}{Contrast Subset} & CW & CH & TP\&TN & \textbf{Total} & \textbf{Decline}\\
 & & & \textit{Acc (\%)} & \textit{Acc (\%)} & \textit{Acc (\%)} & \textbf{Acc (\%)} & \textbf{(\%)} \\
\midrule
\midrule

\rowcolor{gray!10} \textbf{1} & \textbf{Base Model} & \textbf{- (Main Set)} & \textbf{51.92} & \textbf{56.27} & \textbf{47.05} & \textbf{51.45} & \textbf{-} \\
 & & \hspace{3mm} $\mathrm{Verb}_{\mathrm{high}}$ / $\mathrm{Verb}_{\mathrm{low}}$ & 45.84 / 49.28 & 50.70 / 52.92 & 40.95 / 42.48 & 45.45 / 48.22 & -11.66 / -6.28 \\
 & & \hspace{3mm} $\mathrm{Noun}_{\mathrm{high}}$ / $\mathrm{Noun}_{\mathrm{low}}$ & 49.36 / 49.28 & 54.04 / 54.87 & 42.48 / 44.95 & 48.45 / 49.16 & -5.83 / -4.45 \\
 & & \hspace{3mm} $\mathrm{Gender}$ & 52.16 & 54.87 & 45.90 & 51.08 & -0.72 \\
 & & \hspace{3mm} \textit{Average} & 49.16 & 53.13 & 42.95 & 48.30 & -6.12 \\
\hline

\rowcolor{gray!10} \textbf{4} & \textbf{Ours w/o LLM w/o Cap \cite{Li2023IntentQA}} & \textbf{- (Main Set)} & \textbf{55.28} & \textbf{61.56} & \textbf{47.81} & \textbf{54.50} & \textbf{-} \\
 & & \hspace{3mm} $\mathrm{Verb}_{\mathrm{high}}$ / $\mathrm{Verb}_{\mathrm{low}}$ & 47.60 / 52.08 & 53.48 / 58.50 & 40.57 / 45.14 & 46.86 / 51.45 & -14.02 / -5.60 \\
 & & \hspace{3mm} $\mathrm{Noun}_{\mathrm{high}}$ / $\mathrm{Noun}_{\mathrm{low}}$ & 52.16 / 52.00 & 58.22 / 59.61 & 43.81 / 45.90 & 51.12 / 51.78 & -6.20 / -4.99 \\
 & & \hspace{3mm} $\mathrm{Gender}$ & 54.72 & 60.45 & 47.05 & 53.80 & -1.28 \\
 & & \hspace{3mm} \textit{Average} & 51.71 & 58.05 & 44.49 & 51.00 & -6.42 \\
\hline

\rowcolor{gray!10} \textbf{9} & \textbf{Qwen \cite{Qwen-VL}} & \textbf{- (Main Set)} & \textbf{58.96} & \textbf{58.77} & \textbf{46.67} & \textbf{55.90} & \textbf{-} \\
 & & \hspace{3mm} $\mathrm{Verb}_{\mathrm{high}}$ / $\mathrm{Verb}_{\mathrm{low}}$ & 53.76 / 55.92 & 49.03 / 55.15 & 43.05 / 44.95 & 50.33 / 53.09 & -9.96 / -5.03 \\
 & & \hspace{3mm} $\mathrm{Noun}_{\mathrm{high}}$ / $\mathrm{Noun}_{\mathrm{low}}$ & 55.20 / 56.16 & 52.92 / 56.55 & 44.95 / 44.76 & 52.30 / 53.42 & -6.44 / -4.44 \\
 & & \hspace{3mm} $\mathrm{Gender}$ & 58.16 & 57.94 & 44.76 & 54.83 & -1.91 \\
 & & \hspace{3mm} \textit{Average} & 55.84 & 54.32 & 44.49 & 52.79 & -5.56 \\
\hline

\rowcolor{gray!10} \textbf{0} & \textbf{Blind LLM (InstructGPT)} & \textbf{- (Main Set)} & \textbf{52.16} & \textbf{62.28} & \textbf{43.43} & \textbf{51.55} & \textbf{-} \\
 & & \hspace{3mm} $\mathrm{Verb}_{\mathrm{high}}$ / $\mathrm{Verb}_{\mathrm{low}}$ & 51.12 / 52.24 & 57.38 / 58.22 & 44.95 / 45.33 & 50.66 / 51.55 & -1.73 / 0.00 \\
 & & \hspace{3mm} $\mathrm{Noun}_{\mathrm{high}}$ / $\mathrm{Noun}_{\mathrm{low}}$ & 50.56 / 51.92 & 58.22 / 58.50 & 40.76 / 41.33 & 49.44 / 50.42 & -4.09 / -2.19 \\
 & & \hspace{3mm} $\mathrm{Gender}$ & 54.64 & 64.07 & 44.76 & 53.80 & +4.36 \\
 & & \hspace{3mm} \textit{Average} & 52.10 & 59.28 & 43.43 & 51.17 & -0.73 \\
\hline

\rowcolor{gray!10} \textbf{0-1} & \textbf{Blind LLM (GPT-4)} & \textbf{- (Main Set)} & \textbf{55.20} & \textbf{63.23} & \textbf{47.24} & \textbf{54.59} & \textbf{-} \\
 & & \hspace{3mm} $\mathrm{Verb}_{\mathrm{high}}$ / $\mathrm{Verb}_{\mathrm{low}}$ & 56.32 / 58.56 & 57.94 / 62.12 & 45.90 / 49.52 & 54.03 / 56.04 & -1.03 / +2.66 \\
 & & \hspace{3mm} $\mathrm{Noun}_{\mathrm{high}}$ / $\mathrm{Noun}_{\mathrm{low}}$ & 53.60 / 56.40 & 62.40 / 62.40 & 46.48 / 46.67 & 53.33 / 55.01 & -2.31 / +0.77 \\
 & & \hspace{3mm} $\mathrm{Gender}$ & 58.56 & 65.74 & 47.81 & 57.12 & +4.63 \\
 & & \hspace{3mm} \textit{Average} & 56.69 & 62.12 & 47.28 & 55.11 & +0.94 \\
\hline

\rowcolor{gray!10} \textcolor{black}{\textbf{7}} & \textcolor{black}{\textbf{Ours Full (InstructGPT)}} & \textbf{- (Main Set)} & \textbf{60.95} & \textbf{67.69} & \textbf{55.43} & \textbf{60.73} & \textbf{-} \\
 & & \hspace{3mm} $\mathrm{Verb}_{\mathrm{high}}$ / $\mathrm{Verb}_{\mathrm{low}}$ & 54.00 / 58.24 & 59.33 / 66.57 & 48.19 / 52.00 & 53.47 / 58.11 & -11.95 / -4.31 \\
 & & \hspace{3mm} $\mathrm{Entity}_{\mathrm{high}}$ / $\mathrm{Entity}_{\mathrm{low}}$ & 56.80 / 58.32 & 66.30 / 66.85 & 52.19 / 53.33 & 57.26 / 58.53 & -5.71 / -3.62 \\
 & & \hspace{3mm} $\mathrm{Gender}$ & 61.36 & 68.52 & 55.43 & 61.11 & +0.63 \\
 & & \hspace{3mm} \textit{Average} & 57.74 & 65.51 & 52.23 & 57.70 & -4.99 \\
\hline

\rowcolor{gray!10} \textcolor{black}{\textbf{7-1}} & \textcolor{black}{\textbf{Ours Full (ChatGPT)}} & \textbf{- (Main Set)} & \textbf{59.28} & \textbf{67.97} & \textbf{53.33} & \textbf{59.28} & \textbf{-} \\
 & & \hspace{3mm} $\mathrm{Verb}_{\mathrm{high}}$ / $\mathrm{Verb}_{\mathrm{low}}$ & 51.92 / 56.24 & 61.28 / 67.13 & 48.57 / 48.57 & 52.67 / 56.19 & -11.15 / -5.21 \\
 & & \hspace{3mm} $\mathrm{Noun}_{\mathrm{high}}$ / $\mathrm{Noun}_{\mathrm{low}}$ & 55.60 / 57.12 & 65.74 / 67.13 & 47.81 / 49.90 & 55.39 / 57.03 & -6.56 / -3.80 \\
 & & \hspace{3mm} $\mathrm{Gender}$ & 60.16 & 68.25 & 51.62 & 59.42 & +0.24 \\
 & & \hspace{3mm} \textit{Average} & 56.21 & 65.91 & 49.29 & 56.14 & -5.30 \\
\hline

\rowcolor{gray!10} \textcolor{black}{\textbf{7-2}} & \textcolor{black}{\textbf{Ours Full (GPT-4)}} & \textbf{- (Main Set)} & \textbf{69.90} & \textbf{68.80} & \textbf{50.67} & \textbf{64.81} & \textbf{-} \\
 & & \hspace{3mm} $\mathrm{Verb}_{\mathrm{high}}$ / $\mathrm{Verb}_{\mathrm{low}}$ & 63.84 / 67.28 & 62.95 / 66.02 & 47.43 / 49.52 & 59.65 / 62.70 & -7.96 / -3.26 \\
 & & \hspace{3mm} $\mathrm{Noun}_{\mathrm{high}}$ / $\mathrm{Noun}_{\mathrm{low}}$ & 65.36 / 67.44 & 66.57 / 67.13 & 50.10 / 50.10 & 61.81 / 63.12 & -4.63 / -2.61 \\
 & & \hspace{3mm} $\mathrm{Gender}$ & 69.20 & 68.52 & 51.05 & 64.62 & -0.29 \\
 & & \hspace{3mm} \textit{Average} & 66.62 & 66.24 & 49.64 & 62.38 & -3.75 \\
\hline

\rowcolor{gray!10} \textbf{-} & \textbf{Human} & \textbf{- (Main Set)} & \textbf{77.76} & \textbf{80.22} & \textbf{79.05} & \textbf{78.49} & \textbf{-} \\
 & & \hspace{3mm} $\mathrm{Verb}_{\mathrm{high}}$ & 76.08 & 76.88 & 75.81 & 76.15 & -2.98 \\
\bottomrule
\end{tabular}}}
\end{center}
\vspace{-0.2cm}
\caption{Comparison results with other representative methods on the Contrast Set. \textcolor{black}{The table is compacted by displaying contrast subsets in paired rows (e.g., High / Low Similarity) while retaining all specific data points with 2 decimal precision.} Model IDs are consistent with the text (Model 7 is the Full Model).}
\label{tab:main result on contrast set}
\end{table*}

\begin{table*}[!tbp]
\begin{center}
\scriptsize{
\resizebox{\textwidth}{!}{%
\begin{tabular}{cllccccc}
\toprule
 \multirow{2}{*}{ID} & \multirow{2}{*}{Model Variant} & \multirow{2}{*}{Contrast Subset} & CW & CH & TP\&TN & \textbf{Total} & \textbf{Decline}\\
 & & & \textit{Acc (\%)} & \textit{Acc (\%)} & \textit{Acc (\%)} & \textbf{Acc (\%)} & \textbf{(\%)} \\
\midrule
\midrule

\rowcolor{gray!10} \textcolor{black}{\textbf{7}} & \textcolor{black}{\textbf{Ours Full (InstructGPT)}} & \textbf{- (Main Set)} & \textbf{60.95} & \textbf{67.69} & \textbf{55.43} & \textbf{60.73} & \textbf{-} \\
 & & \hspace{3mm} $\mathrm{Verb}_{\mathrm{high}}$ / $\mathrm{Verb}_{\mathrm{low}}$ & 54.00 / 58.24 & 59.33 / 66.57 & 48.19 / 52.00 & 53.47 / 58.11 & -11.95 / -4.31 \\
 & & \hspace{3mm} $\mathrm{Entity}_{\mathrm{high}}$ / $\mathrm{Entity}_{\mathrm{low}}$ & 56.80 / 58.32 & 66.30 / 66.85 & 52.19 / 53.33 & 57.26 / 58.53 & -5.71 / -3.62 \\
 & & \hspace{3mm} $\mathrm{Gender}$ & 61.36 & 68.52 & 55.43 & 61.11 & +0.63 \\
 & & \hspace{3mm} \textit{Average} & 57.74 & 65.51 & 52.23 & 57.70 & -4.99 \\
\hline

\rowcolor{gray!10} \textcolor{black}{\textbf{7-1}} & \textcolor{black}{\textbf{Ours Full (ChatGPT)}} & \textbf{- (Main Set)} & \textbf{59.28} & \textbf{67.97} & \textbf{53.33} & \textbf{59.28} & \textbf{-} \\
 & & \hspace{3mm} $\mathrm{Verb}_{\mathrm{high}}$ / $\mathrm{Verb}_{\mathrm{low}}$ & 51.92 / 56.24 & 61.28 / 67.13 & 48.57 / 48.57 & 52.67 / 56.19 & -11.15 / -5.21 \\
 & & \hspace{3mm} $\mathrm{Noun}_{\mathrm{high}}$ / $\mathrm{Noun}_{\mathrm{low}}$ & 55.60 / 57.12 & 65.74 / 67.13 & 47.81 / 49.90 & 55.39 / 57.03 & -6.56 / -3.80 \\
 & & \hspace{3mm} $\mathrm{Gender}$ & 60.16 & 68.25 & 51.62 & 59.42 & +0.24 \\
 & & \hspace{3mm} \textit{Average} & 56.21 & 65.91 & 49.29 & 56.14 & -5.30 \\
\hline

\rowcolor{gray!10} \textcolor{black}{\textbf{7-2}} & \textcolor{black}{\textbf{Ours Full (GPT-4)}} & \textbf{- (Main Set)} & \textbf{69.90} & \textbf{68.80} & \textbf{50.67} & \textbf{64.81} & \textbf{-} \\
 & & \hspace{3mm} $\mathrm{Verb}_{\mathrm{high}}$ / $\mathrm{Verb}_{\mathrm{low}}$ & 63.84 / 67.28 & 62.95 / 66.02 & 47.43 / 49.52 & 59.65 / 62.70 & -7.96 / -3.26 \\
 & & \hspace{3mm} $\mathrm{Noun}_{\mathrm{high}}$ / $\mathrm{Noun}_{\mathrm{low}}$ & 65.36 / 67.44 & 66.57 / 67.13 & 50.10 / 50.10 & 61.81 / 63.12 & -4.63 / -2.61 \\
 & & \hspace{3mm} $\mathrm{Gender}$ & 69.20 & 68.52 & 51.05 & 64.62 & -0.29 \\
 & & \hspace{3mm} \textit{Average} & 66.62 & 66.24 & 49.64 & 62.38 & -3.75 \\
\hline
\hline

\rowcolor{gray!10} \textcolor{black}{\textbf{6}} & \textcolor{black}{\textbf{Ours w/o Caption (InstructGPT)}} & \textbf{- (Main Set)} & \textbf{58.48} & \textbf{66.02} & \textbf{53.33} & \textbf{58.48} & \textbf{-} \\
 & & \hspace{3mm} $\mathrm{Verb}_{\mathrm{high}}$ / $\mathrm{Verb}_{\mathrm{low}}$ & 51.36 / 54.56 & 59.61 / 62.67 & 45.71 / 51.05 & 51.36 / 55.06 & -12.18 / -5.85 \\
 & & \hspace{3mm} $\mathrm{Noun}_{\mathrm{high}}$ / $\mathrm{Noun}_{\mathrm{low}}$ & 55.12 / 55.52 & 65.46 / 65.18 & 48.95 / 50.29 & 55.34 / 55.86 & -5.37 / -4.48 \\
 & & \hspace{3mm} $\mathrm{Gender}$ & 58.80 & 66.57 & 52.76 & 58.62 & +0.24 \\
 & & \hspace{3mm} \textit{Average} & 55.07 & 63.90 & 49.75 & 55.25 & -5.53 \\
\hline

\rowcolor{gray!10} \textcolor{black}{\textbf{6-1}} & \textcolor{black}{\textbf{Ours w/o Caption (ChatGPT)}} & \textbf{- (Main Set)} & \textbf{57.92} & \textbf{65.46} & \textbf{50.86} & \textbf{57.45} & \textbf{-} \\
 & & \hspace{3mm} $\mathrm{Verb}_{\mathrm{high}}$ / $\mathrm{Verb}_{\mathrm{low}}$ & 50.16 / 54.88 & 55.99 / 61.84 & 45.33 / 46.86 & 49.95 / 54.08 & -13.05 / -5.87 \\
 & & \hspace{3mm} $\mathrm{Noun}_{\mathrm{high}}$ / $\mathrm{Noun}_{\mathrm{low}}$ & 52.40 / 53.28 & 60.72 / 62.67 & 44.57 / 47.05 & 51.87 / 53.33 & -9.71 / -7.17 \\
 & & \hspace{3mm} $\mathrm{Gender}$ & 57.60 & 64.62 & 49.33 & 56.75 & -1.22 \\
 & & \hspace{3mm} \textit{Average} & 53.66 & 61.17 & 46.63 & 53.20 & -7.40 \\
\hline

\rowcolor{gray!10} \textcolor{black}{\textbf{6-2}} & \textcolor{black}{\textbf{Ours w/o Caption (GPT-4)}} & \textbf{- (Main Set)} & \textbf{63.28} & \textbf{72.98} & \textbf{54.10} & \textbf{62.65} & \textbf{-} \\
 & & \hspace{3mm} $\mathrm{Verb}_{\mathrm{high}}$ / $\mathrm{Verb}_{\mathrm{low}}$ & 59.04 / 60.56 & 63.79 / 70.19 & 51.62 / 52.95 & 58.01 / 60.31 & -7.41 / -3.74 \\
 & & \hspace{3mm} $\mathrm{Noun}_{\mathrm{high}}$ / $\mathrm{Noun}_{\mathrm{low}}$ & 59.92 / 60.40 & 68.80 / 68.52 & 52.38 / 51.81 & 59.56 / 59.65 & -4.93 / -4.79 \\
 & & \hspace{3mm} $\mathrm{Gender}$ & 62.88 & 72.98 & 55.24 & 62.70 & +0.07 \\
 & & \hspace{3mm} \textit{Average} & 60.56 & 68.86 & 52.80 & 60.05 & -4.16 \\
\hline
\hline

\rowcolor{gray!10} \textbf{8} & \textbf{Caption + InstructGPT} & \textbf{- (Main Set)} & \textbf{56.56} & \textbf{58.22} & \textbf{45.14} & \textbf{54.03} & \textbf{-} \\
 & & \hspace{3mm} $\mathrm{Verb}_{\mathrm{high}}$ / $\mathrm{Verb}_{\mathrm{low}}$ & 54.80 / 56.49 & 53.20 / 55.99 & 47.62 / 46.67 & 52.76 / 53.94 & -2.35 / -0.17 \\
 & & \hspace{3mm} $\mathrm{Entity}_{\mathrm{high}}$ / $\mathrm{Entity}_{\mathrm{low}}$ & 53.60 / 55.60 & 55.71 / 56.82 & 47.43 / 45.52 & 52.44 / 53.33 & -2.94 / -1.30 \\
 & & \hspace{3mm} $\mathrm{Gender}$ & 57.04 & 60.45 & 48.38 & 55.48 & +2.68 \\
 & & \hspace{3mm} \textit{Average} & 55.51 & 56.43 & 47.12 & 53.59 & -0.82 \\
\hline

\rowcolor{gray!10} \textbf{8-1} & \textbf{Caption + ChatGPT} & \textbf{- (Main Set)} & \textbf{57.36} & \textbf{64.35} & \textbf{44.95} & \textbf{55.48} & \textbf{-} \\
 & & \hspace{3mm} $\mathrm{Verb}_{\mathrm{high}}$ / $\mathrm{Verb}_{\mathrm{low}}$ & 53.68 / 59.96 & 55.43 / 61.56 & 43.43 / 46.29 & 51.45 / 55.11 & -7.26 / -0.67 \\
 & & \hspace{3mm} $\mathrm{Noun}_{\mathrm{high}}$ / $\mathrm{Noun}_{\mathrm{low}}$ & 53.04 / 56.16 & 60.45 / 61.00 & 44.19 / 44.76 & 52.11 / 54.17 & -6.07 / -2.36 \\
 & & \hspace{3mm} $\mathrm{Gender}$ & 57.84 & 65.18 & 46.67 & 56.33 & +1.53 \\
 & & \hspace{3mm} \textit{Average} & 56.14 & 60.72 & 45.07 & 53.83 & -2.97 \\
\hline

\rowcolor{gray!10} \textbf{8-2} & \textbf{Caption + GPT-4} & \textbf{- (Main Set)} & \textbf{68.16} & \textbf{67.13} & \textbf{48.76} & \textbf{63.21} & \textbf{-} \\
 & & \hspace{3mm} $\mathrm{Verb}_{\mathrm{high}}$ / $\mathrm{Verb}_{\mathrm{low}}$ & 64.32 / 65.44 & 55.99 / 62.95 & 45.14 / 49.33 & 58.20 / 61.06 & -7.93 / -3.40 \\
 & & \hspace{3mm} $\mathrm{Noun}_{\mathrm{high}}$ / $\mathrm{Noun}_{\mathrm{low}}$ & 63.52 / 66.00 & 67.41 / 65.74 & 49.52 / 49.14 & 60.73 / 61.81 & -3.92 / -2.21 \\
 & & \hspace{3mm} $\mathrm{Gender}$ & 66.24 & 68.80 & 49.52 & 62.56 & -1.03 \\
 & & \hspace{3mm} \textit{Average} & 65.10 & 64.18 & 48.53 & 60.87 & -3.70 \\

\bottomrule
\end{tabular}}}
\end{center}
\vspace{-0.2cm}
\caption{Ablation analysis on integrating LLM for common sense reasoning (Contrast Set). \textcolor{black}{All data points are preserved using the paired-row format (High / Low) for compactness.} Model IDs: 7 (Full), 6 (w/o Caption), 8 (Caption only).}
\label{tab:Ablation Contrast}
\end{table*}

\begin{figure*}[!tbp]
\begin{center}
   \includegraphics[width=0.8\textwidth]{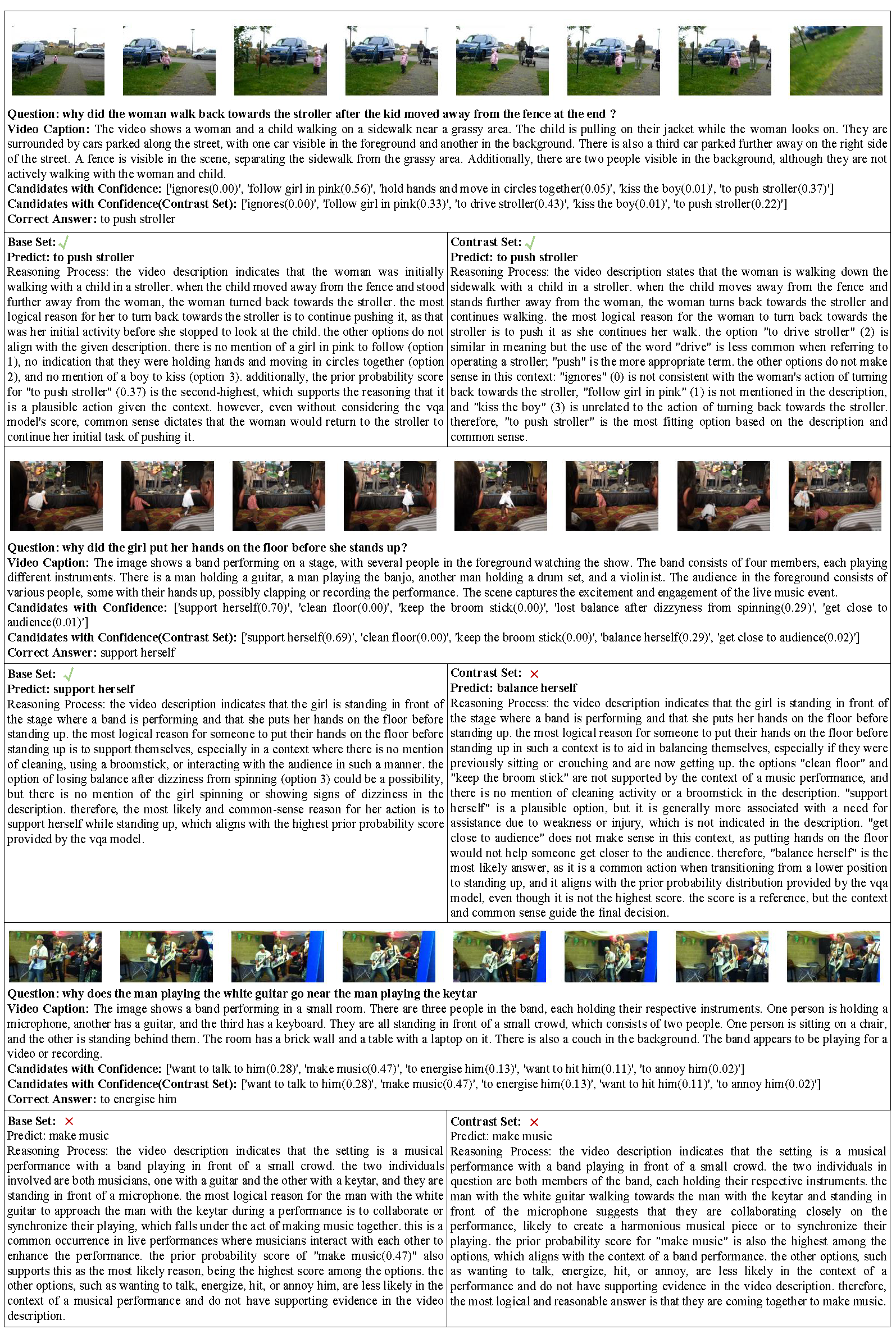}
\end{center}
   \caption{Qualitative analysis and failure cases of ours model (with GPT-4) on contrast sets.}
\label{fig:visualize_contrast_GPT4}
\end{figure*}

\subsection{Experiments on the Contrast Set}
\label{Sec:Experiments on the Contrast Set}

\subsubsection{\textbf{How Effective is the Contrast Set?}}
\label{sec:The Effectiveness of the Contrast Set}
We verify the effectiveness of our contrast set in evaluating models and use ``Decline,'' i.e., the relative accuracy decline percentage, as the metric to measure model performance robustness. This ``Decline'' is referred to as the Contrast Performance Decline, and the formula is given as follows:

\begin{equation}
\text{Contrast Performance Decline} = \left( \frac{A_{\text{base}} - A_{\text{contrast}}}{A_{\text{base}}} \right) \times 100\%,
\end{equation}
where \( A_{\text{base}} \) represents the accuracy on the main set, and \( A_{\text{contrast}} \) represents the accuracy on the contrast set. The formula calculates the percentage decrease in accuracy from the main set to the contrast set, providing a measure of how much the model's performance declines on the contrast set relative to its performance on the main set.

As shown in \Cref{tab:main result on contrast set}, ``Ours w/o LLM w/o Caption'' (Model 4) and the current advanced vision-language model ``Qwen'' \cite{Qwen-VL} all demonstrated a notable performance decrease on the contrast set, especially when faced with verbs of higher similarity (i.e., substituting an verb phrase in the contrast set that fits the video scene but is incorrect answer to the question). Humans also demonstrated a performance decline on the ``$Verb_{high}$'' subset of the contrast set, but only by 2.98\%, whereas traditional models (e.g., Model 1, 4 and 9 in \Cref{tab:main result on contrast set}) generally declined by more than 10\%. One example is from the high-similarity verb contrast set (\Cref{fig:visualize_contrast_GPT4}). In this example, the question is ``Why did the woman walk back towards the stroller after the kid moved away from the fence at the end?'' The ground truth option is ``to push the stroller,'' while the contrast option is ``to drive the stroller.'' The reason for choosing the ground truth option is that after the child moved away, the woman needed to push the stroller to catch up with the child, implying that they were leaving the area. These two answers are semantically very similar (controlled during dataset generation) and somewhat match the video features (a stroller being moved). This makes the question more challenging for traditional models, which rely on pattern matching, but simpler for models using reasoning. A reasoning-based model would only need to use common sense to realize that strollers are typically pushed rather than driven by a driver, thus selecting the correct option. The reason why traditional models show much more accuracy decline might be that they rely more on just matching options to the video based on surface patterns rather than truly deriving answers through reasoning. This makes them severely affected by the contrast set.

We further experimented with the ``Blind LLM'' approach (i.e., only InstructGPT or GPT-4) on the contrast set, letting the LLM decide the most appropriate option based solely on the question. We observed that although the Blind LLM approach did not achieve high performance, it exhibited good robustness on the contrast set, achieving a similar accuracy decline as humans. This suggests that reasoning through commonsense is closer to how humans tackle intent-based QA problems, as opposed to the traditional models' approach of simply ``retrieving'' and ``matching'' answers based on surface patterns. These series of experiments also demonstrate the effectiveness of our contrast set in evaluating a model's robustness on the IntentQA task.

\subsubsection{\textbf{How Does Our Approach Leverage LLMs to Enhance Performance and Robustness?}}
\label{sec: ours with llm}

Our method demonstrates two key improvements: 1) enhanced performance across all sets, including the main set and contrast subsets, and 2) improved robustness compared to the Traditional V-L model.

First, our method outperforms both the Blind LLM and the Traditional V-L model on the main set and all five contrast subsets. As shown in the ``Ours'' section of \Cref{tab:main result on contrast set}, our full model's performance on the main set improved from 54.50\% to \textcolor{black}{60.73\% with InstructGPT (Model 7)}, to \textcolor{black}{59.28\% with ChatGPT (Model 7-1)}, and to \textcolor{black}{64.81\% with GPT-4 (Model 7-2)}, demonstrating significant performance enhancement.

Compared to methods using pure language models without visual input, our approach also shows marked improvements. As shown in the ``Blind LLM'' section of \Cref{tab:main result on contrast set}, our method improves performance from 51.55\% to \textcolor{black}{60.73\%} over InstructGPT alone, and from 54.59\% to \textcolor{black}{64.81\%} over GPT-4 alone.

In terms of robustness, while our method shows some disadvantages compared to language models alone (as shown in the ``Blind LLM" section of \Cref{tab:main result on contrast set}), it significantly outperforms the Traditional V-L model. Model 1 and Model 4 exhibit substantial performance decline in the $\mathrm{Verb}_{\mathrm{high}}$ set and average declines of over 6\% across the five contrast subsets (6.12\% for Model 1 and 6.42\% for Model 4). The large-scale V-L model Qwen shows a 5.56\% average decline. In contrast, the \textcolor{black}{Model 7 series} in the ``Ours" section shows smaller declines: 5.3\% with ChatGPT and only 3.75\% with GPT-4.

\subsubsection{\textbf{How Do VideoQA Likelihood and Caption Affect Our Full Model in the Contrast Set?}}

To further investigate the impact of VideoQA likelihood and captions on our model's performance, we conducted experiments on five contrast subsets, as summarized in \Cref{tab:Ablation Contrast}. These experiments aim to understand how these information channels contribute to the model's performance and robustness in different contexts.

From the experiments detailed in \Cref{tab:commonsense reasoning analysis}, we observed that removing either the VideoQA likelihood (Model 8) or the captions (\textcolor{black}{Model 6}) leads to a decrease in performance. Notably, the absence of VideoQA likelihood resulted in a more substantial decline compared to the removal of captions, highlighting its significant role in enhancing model performance. This trend persists in the contrast set experiments, where we analyzed the average decline across five different subsets. The contrast set analysis revealed that while the VideoQA likelihood significantly boosts model performance, it also introduces fluctuations in performance, thus affecting the model's robustness. Specifically, the average decline in performance for Model 8, which excludes the VideoQA likelihood, is 0.82\%, 2.97\%, and 3.70\% across different subsets. In comparison, \textcolor{black}{Model 6}, which excludes captions, shows a higher average decline of 5.53\%, 7.40\%, and 4.16\%. These results indicate that reliance on VideoQA likelihood improves performance but at the cost of increased sensitivity to changes in input data.

Interestingly, using a more advanced reasoning model, such as GPT-4, mitigates the performance drop caused by VideoQA likelihood. This suggests that enhancing the model's inherent reasoning capabilities can counterbalance the dependence on external information channels. Furthermore, incorporating captions consistently enhances robustness, as evidenced by the average declines of 4.99\%, 5.30\%, and 3.75\% in \textcolor{black}{Model 7}, which includes captions but varies other components. This shows that while captions are not as impactful as VideoQA likelihood in boosting overall performance, they contribute to a more stable and reliable model output.

In summary, the combination of VideoQA likelihood and captions strikes a balance between achieving high performance and maintaining robustness. While VideoQA likelihood significantly enhances performance, it also increases the model's sensitivity to changes, which can be mitigated by using models with stronger reasoning capabilities like GPT-4. Captions, on the other hand, offer consistent support in maintaining robustness, making them a valuable component in the overall model design.

\subsubsection{\textbf{Can Using LLM as a Reasoner Really Solve Problems?}}

\Cref{fig:visualize_contrast_GPT4} showcases examples of the reasoning processes of some of our best models, including both successes and failures on the base and contrast sets. These examples help us qualitatively analyze the characteristics of LLM reasoning and identify current methodological issues.

In the first example in \Cref{fig:visualize_contrast_GPT4}, our method correctly answered the question on both the base and contrast sets. The video question is: ``Why did the woman walk back towards the stroller after the kid moved away from the fence at the end?'' The correct answer is that the woman returned to ``push the stroller,'' whereas the VQA likelihood model chose ``follow the girl in pink.'' Both options match the video content, but ``push the stroller'' answers the question correctly. The VQA likelihood model tends to select the option that best matches the video content, rather than reasoning to find the correct answer. The LLM’s reasoning process, shown in the left half of the first example in \Cref{fig:visualize_contrast_GPT4}, demonstrates the correct reasoning. The LLM first analyzes the scene described in the caption, infers that the woman would push the stroller using common sense, and then eliminates other options. For the contrast option ``to drive the stroller,'' the LLM rules it out by reasoning that ``drive'' is an uncommon term for operating a stroller. However, the VQA likelihood model struggles here; it incorrectly rated ``to drive the stroller'' higher than ``to push the stroller'' due to semantic similarities, without considering common sense. This reasoning approach is also evident in other examples, regardless of correctness. In the second-row example, although our method failed on the contrast set, it still followed the process of scene analysis, common-sense reasoning, and eliminating other options.

We now analyze these failures to understand their causes. In the second-row example, the question is: ``Why did the girl put her hands on the floor before she stands up?'' In the video, the girl falls and uses her hands to ``support herself.'' Our method incorrectly chose ``balance herself'' on the contrast set. The LLM’s reasoning was not flawed; the girl might have needed to ``balance herself'' if performing a low-position move. However, the LLM missed the detail of the girl falling, which was not evident in the caption. Thus, while LLMs can significantly address the VQA likelihood model's focus on option-video matching, they cannot resolve the issue of missing critical video details.

In the third example, the question is: ``Why does the man playing the white guitar go near the man playing the keytar?'' The correct answer is that he approached to ``energize him'' by playing together. The LLM only identified the superficial aspect of ``making music together,'' missing the deeper intent of mutual motivation. This shows that while our method improves model performance, there remains a significant gap compared to human reasoning. We discuss more failure examples in the supplementary materials, analyzing and categorizing the reasons for these failures.

\section{Conclusion}

\textcolor{black}{In this work, we address the challenge of inferring latent human intentions in videos by introducing the \textbf{IntentQA} task and dataset, grounded in situational, contrastive, and commonsense contexts. 
To ensure a rigorous evaluation beyond simple accuracy, we construct a comprehensive robustness benchmark comprising five distinct \textbf{Contrast Sets}, effectively mitigating dataset biases and revealing true model stability.
Furthermore, we propose \textbf{X-CaVIR}, a novel framework that synergizes visual observations with the reasoning capabilities of Large Language Models (e.g., GPT-4).
By integrating video captions and VQA confidence scores, our pipeline not only achieves state-of-the-art performance but also provides transparent, interpretable reasoning pathways. 
We hope this work serves as a solid step toward building AI systems capable of robust and explainable social intelligence.}

\noindent{\textbf{Acknowledgement}} This research was supported by the National Natural Science Foundation of China (No. U23B2060) and the National Science and Technology Major Project (2022ZD0114900). 

\input{references.tex}

\vspace{-1.2cm}
\begin{IEEEbiography} [{\includegraphics[width=1in,height=1.25in,clip,keepaspectratio]{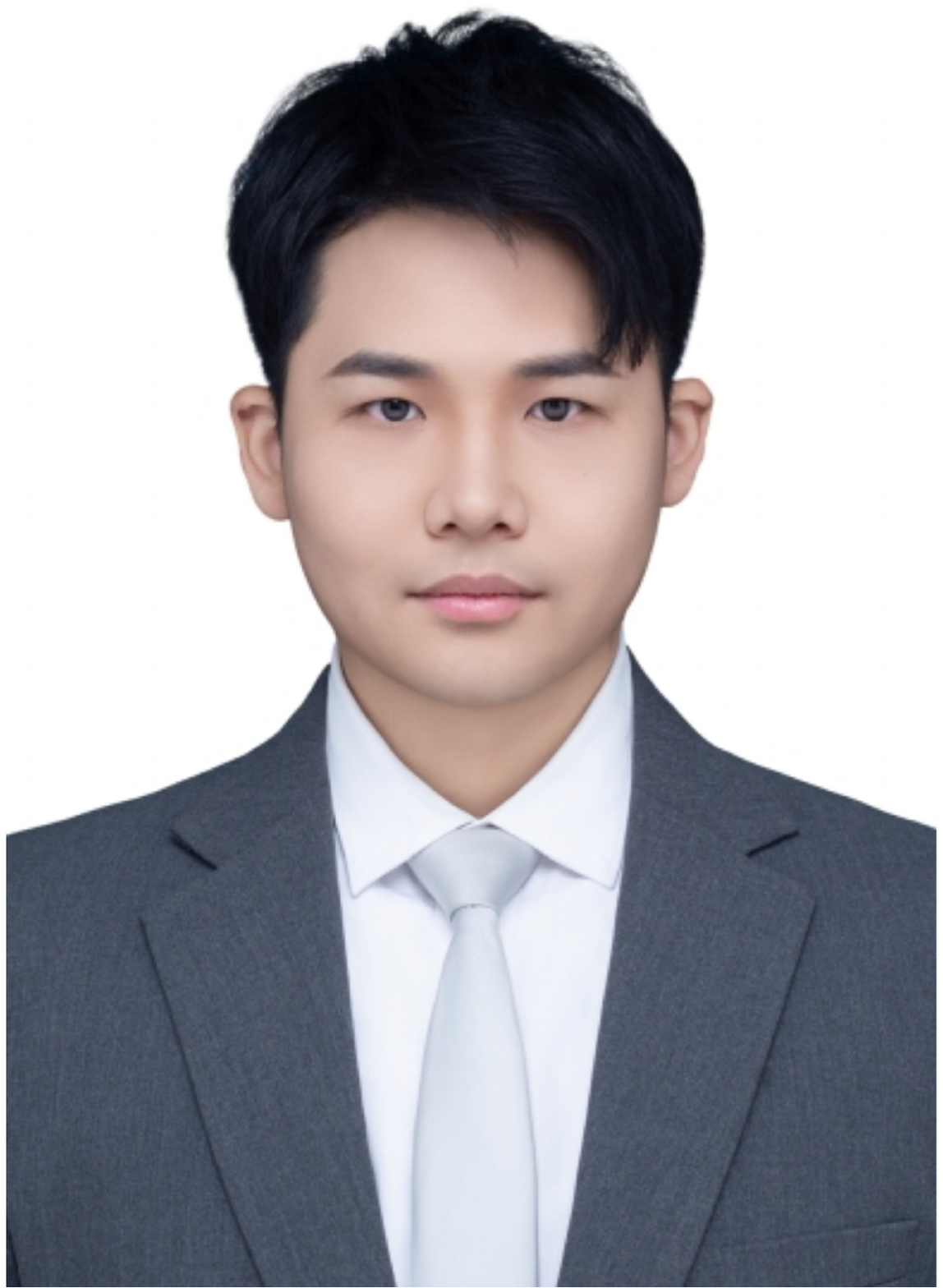}}]{Jiapeng Li} received his B.E. degree from Northeastern University (NEU), China. He is currently pursuing a doctorate in the Institute of Artificial Intelligence and Robotics at Xi’an Jiaotong University (XJTU). Previously, he served as an assistant researcher at the Beijing Institute for General Artificial Intelligence (BIGAI). His research interests include cognitive reasoning, video understanding, and multimodality, etc. 
\end{IEEEbiography}
\vspace{-1.2cm}
\begin{IEEEbiography}[{\includegraphics[width=1in,height=1.25in,clip,keepaspectratio]{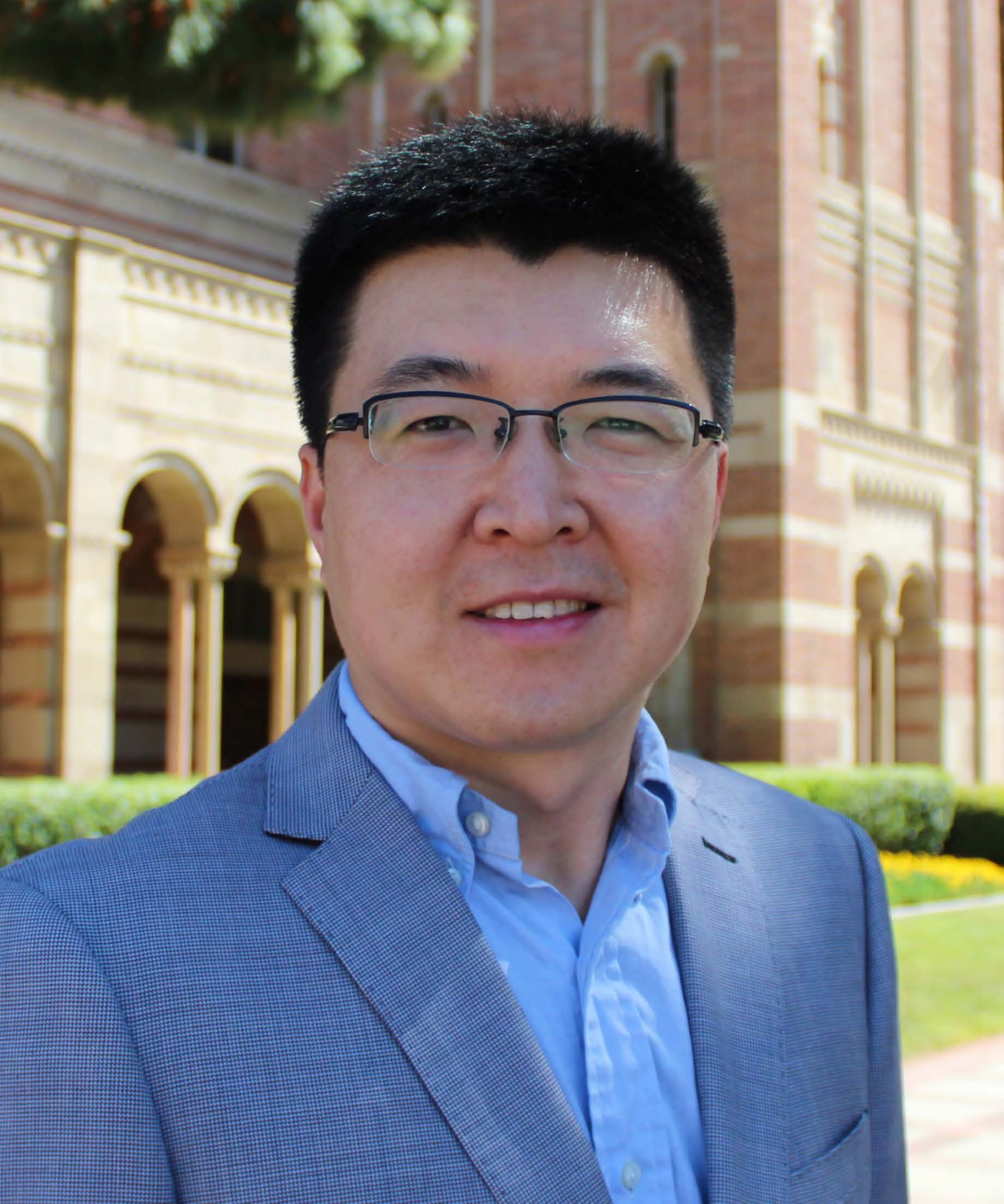}}]{Ping Wei} received the B.E. and Ph.D. degrees from Xi’an Jiaotong University, Xi’an, China. He is currently a Professor with the Institute of Artificial Intelligence and Robotics at Xi’an Jiaotong University. He has been a postdoctoral researcher at University of California, Los Angeles (UCLA). His research interests include computer vision, machine learning, and computational cognition. He is a Senior Member of IEEE.
\end{IEEEbiography}
\vspace{-1.2cm}
\begin{IEEEbiography}
[{\includegraphics[width=1in,height=1.25in,clip,keepaspectratio]{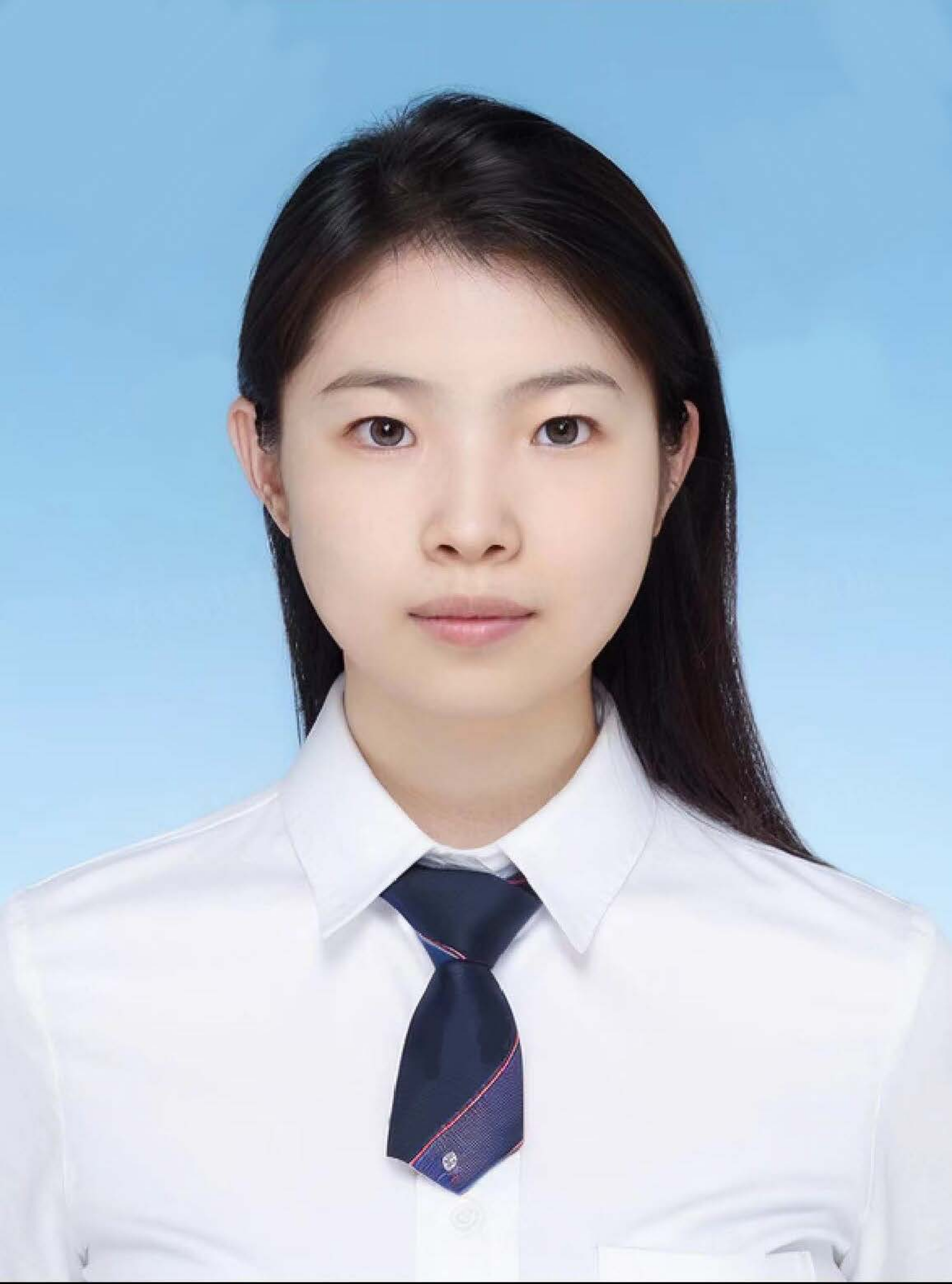}}]{Wenjuan Han} is an associate professor with the School of Information Science and Technology at Beijing Jiaotong University. She used to be a researcher at the National University of Singapore. She received her Ph.D. degree from the joint program of ShanghaiTech University and University of Chinese Academy of Sciences. She was a visiting student at University of California, Los Angeles in 2019. Her research focuses on natural language understanding and multimodality, especially multimodal large language models. Her research has been published in top-tier *ACL conferences. 
\end{IEEEbiography}
\vspace{-1.2cm}
\begin{IEEEbiography}[{\includegraphics[width=1in,height=1.25in,clip,keepaspectratio]{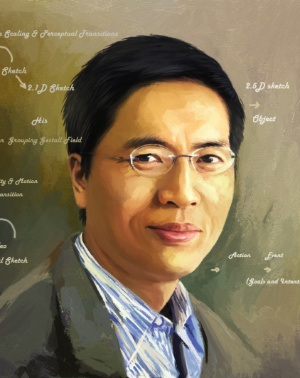}}]{Song-Chun Zhu} (Fellow, IEEE) received the Ph.D. degree from Harvard University, in 1996. He is currently the chair professor jointly with Tsinghua University and Peking University, director of Institute for Artificial Intelligence, Peking University. He worked at Brown, Stanford, Ohio State, and UCLA before returning to China, in 2020, to launch a non-profit organization - Beijing Institute for General Artificial Intelligence. He has published more than 300 papers in computer vision, statistical modeling and learning, cognition, language, robotics, and AI. He received a number of honors, including the Marr Prize, in 2003, the Aggarwal Prize from the Intl Association of Pattern Recognition, in 2008, the Helmholtz Test-of-Time Prize, in 2013, twice Marr Prize Honorary Nominations, in 1999 and 2007, a Sloan Fellowship, the US NSF Career Award, and ONR Young Investigator Award, in 2001. He serves as
a general co-chair for CVPR 2012 and CVPR 2019.
\end{IEEEbiography}
\vspace{-1.2cm}
\begin{IEEEbiography}
[{\includegraphics[width=1in,height=1.25in,clip,keepaspectratio]{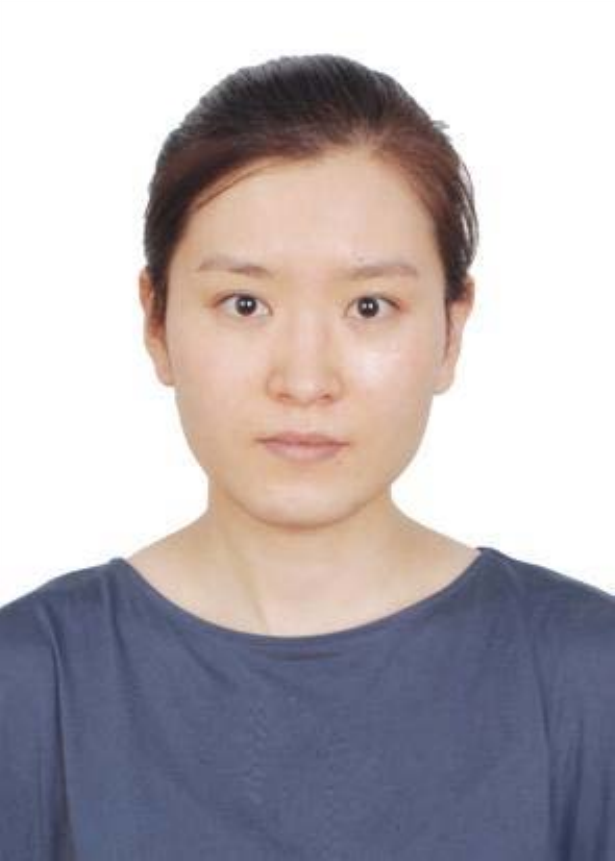}}]{Lifeng Fan} is a research scientist at Beijing Institute for General Artificial Intelligence (BIGAI). She obtained her B.S. degree in Statistics from Zhejiang University at 2016. And she pursued her Ph.D. degree in Statistics from University of California, Los Angeles (UCLA) at 2021 under the supervision of Prof. Song-Chun Zhu. Her research interests include cognitive reasoning, social interaction modeling, Theory of Mind, etc. 
\end{IEEEbiography}

\end{document}